\documentclass{article}
\PassOptionsToPackage{numbers, compress}{natbib}

 \usepackage[main, final]{neurips_2026}

\usepackage[utf8]{inputenc} 
\usepackage[T1]{fontenc}    
\usepackage{hyperref}       
\usepackage{url}            
\usepackage{booktabs}       
\usepackage{amsfonts}       
\usepackage{amsmath}        
\usepackage{nicefrac}       
\usepackage{microtype}      
\usepackage{xcolor}         
\usepackage{graphicx}       
\usepackage{subcaption}     
\usepackage{algorithm}      
\usepackage{algorithmic}    
\usepackage{multirow}       
\usepackage{enumitem}
\usepackage{booktabs}
\usepackage{multirow}
\usepackage{siunitx}

\usepackage{array}
\usepackage{colortbl}
\usepackage{multirow}
\usepackage{graphicx}
\usepackage{xcolor}
\usepackage{amssymb}
\usepackage{pifont}
\usepackage{caption}

\definecolor{binA}{RGB}{70,130,220}
\definecolor{binB}{RGB}{220,100,50}
\definecolor{binC}{RGB}{80,180,80}
\definecolor{selfborder}{RGB}{200,50,50}
\definecolor{allowed}{RGB}{220,240,220}
\definecolor{blocked}{RGB}{240,220,220}

\usepackage{xspace}
\newcommand{\methodname}{MACRO\xspace}

\title{MACRO: Training-free Multi-plane Attention for Closeup Render Optimization}

\author{%
  \textbf{Nitzan Hodos}$^{1,2}$, \textbf{Roy Amoyal}$^{1,3}$, \textbf{Lior Fritz}$^{1}$, \textbf{Ianir Ideses}$^{1}$, \textbf{Sagie Benaim}$^{1,4}$, \textbf{Netalee Efrat}$^{1}$
  \\
  $^{1}$Amazon Prime Video \qquad $^{2}$Technion, Israel Institute of Technology \\
  $^{3}$Ben Gurion University \qquad $^{4}$Hebrew University of Jerusalem
  \\
  \texttt{https://nitzanhod.github.io/MACRO/}
}

\begin{document}

\maketitle

\vspace{-0.1in}
{\centering
  \includegraphics[width=\textwidth, trim=0 0 0 22, clip]{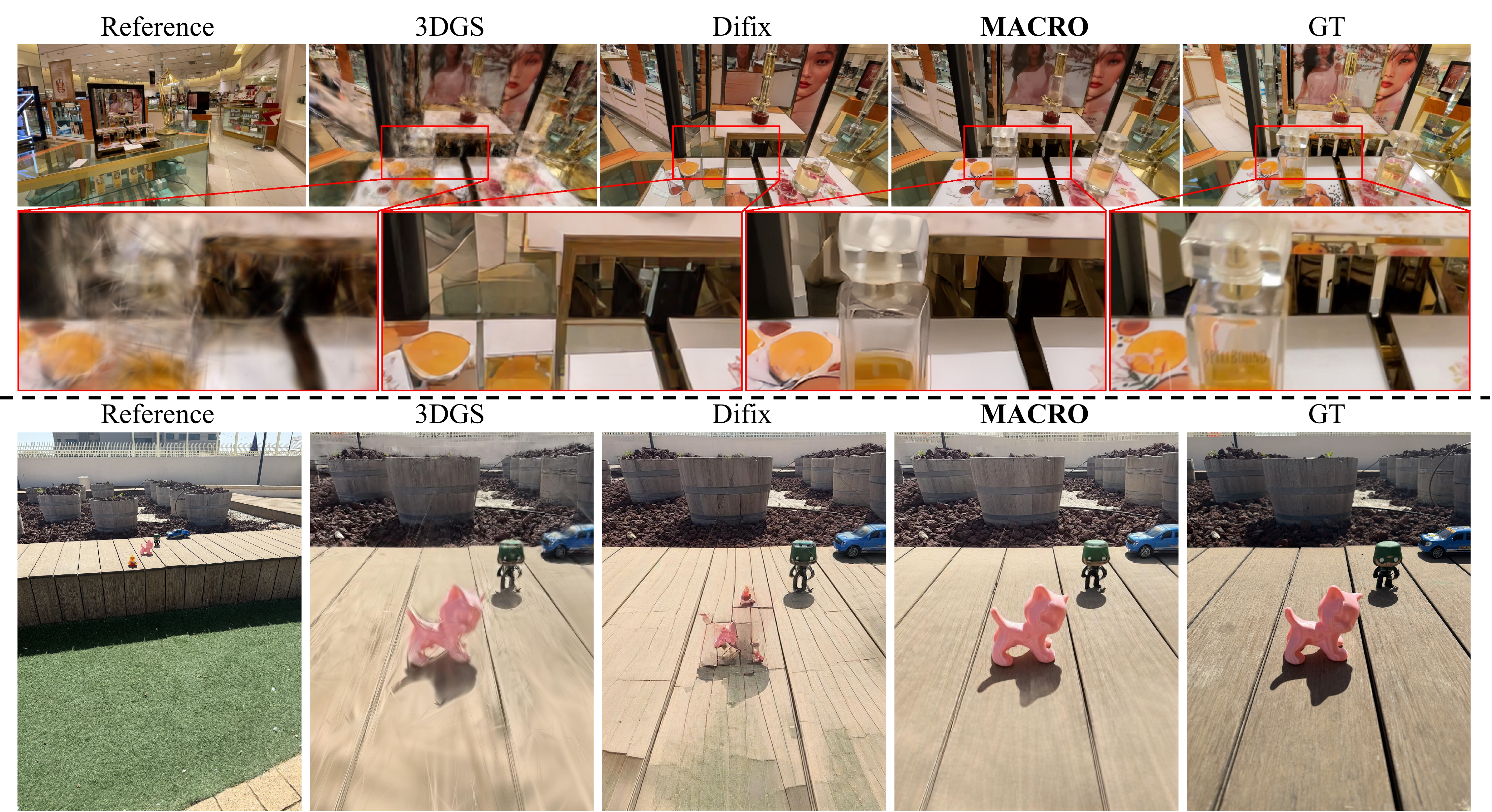}
  \captionof{figure}{\textbf{Close-up novel view synthesis with \methodname{}.} \textbf{Top}: a scene from DL3DV-Closeup. Given a wide reference view, 3DGS produces a blurry close-up rendering. Difix generates a sharp image but transfers incorrect textures from the reference due to scale mismatch (see insets). \methodname{} faithfully restores details that match the ground truth. \textbf{Bottom}: a scene from MobileClose-10. Difix transfers the wood plank texture at the scale of the wide shot, producing visibly wrong patterns. \methodname{} recovers the correct close-up texture without retraining.}
  \label{fig:teaser}
\par}

\begin{abstract}
Close-up rendering, zooming into a scene well beyond any training camera, is important for virtual production and interactive 3D content, yet remains an open challenge. 3D Gaussian splatting (3DGS) enables high-fidelity, real-time novel view synthesis, but its rendering quality degrades at close range. Recent diffusion-based methods that enhance the rendering by conditioning on reference images from the training set produce significant artifacts in this setting. We analyze this failure and identify its root cause: the scale gap between the close-up and reference views. We show that the features in reference-conditioned enhancement models are not scale-invariant, causing cross-view attention to retrieve incorrect correspondences when the same content appears at different scales, and that this mismatch cannot be corrected in latent space because the VAE encoder is not scale-equivariant. Building on this analysis we introduce MACRO, Multi-plane Attention for Closeup Render Optimization, a training-free method for high-quality close-up novel view synthesis from 3DGS. MACRO resolves the scale gap by leveraging the scene's known 3D structure: it decomposes the close-up into depth planes, crops and resizes references in image space to match the scale of each plane before encoding, and applies a depth-aware attention mask so each token attends only to scale-matched references. The method requires no architectural changes or additional training. We further contribute two new close-up novel view synthesis benchmarks, the first standardized evaluation protocol for this setting, and demonstrate state-of-the-art results on both, outperforming existing 3DGS and diffusion-based methods on both reconstruction and perceptual metrics.

\end{abstract}

\section{Introduction}
\label{sec:intro}

3D Gaussian splatting (3DGS)~\cite{kerbl3Dgaussians} has demonstrated impressive fidelity and speed for novel view synthesis. However, rendering quality deteriorates significantly when the synthesized viewpoint deviates from the training views. A particularly important case is close-up view synthesis: rendering zoomed-in views that are substantially closer to the scene than any training camera, often from different viewpoints. Close-up rendering is essential for applications such as virtual production, interactive 3D content, and immersive media, where users expect to freely explore scenes at any distance. This setting is fundamentally different from super-resolution or digital zoom: the close-up view may observe content that is only partially visible across multiple training images, with no single reference providing complete coverage. Moreover, the close-up viewpoint is out-of-distribution, producing severe artifacts in regions with insufficient training signal, while fine details that were unresolved or occluded at the original capture distance cannot be recovered from the 3DGS representation alone.

Since these missing details cannot be recovered from the 3D representation, recent works turn to diffusion-based enhancement methods~\cite{wu2025difix3d,fischer2025flowr,delutio2026artifixer,yin2025gsfixer} that condition on real reference images from the training set. By leveraging the strong image priors of diffusion models together with actual scene observations, these methods can transfer genuine textures and details from the references to the degraded rendering, producing outputs faithful to the true scene. However, in this close-up setting these methods produce poor results. They assume that the target and reference views observe the scene at a similar scale, whereas in close-ups objects appear at completely different scales that vary spatially across the image, as a function of depth. Existing approaches that do target close-up synthesis~\cite{xia2025closeupgs,zhang2026closeupshot} attempt to bridge the gap gradually, generating intermediate views at progressively closer distances. While this progressive strategy reduces the distribution gap at each step, it does not tackle the actual root cause: the objects in the close-up and reference are at different scales. Progressive schemes are also computationally expensive, requiring multiple rounds of rendering, enhancement, and retraining of the 3D representation, with each step introducing accumulated errors.

In this work, we identify two key insights that explain and resolve this failure. First, the VAE and U-Net features used in diffusion-based enhancement are sensitive to scale: when the same scene content is encoded at different scales, the resulting latent representations differ, causing attention between the target and reference tokens to fail. Second, this scale mismatch cannot be corrected in latent space because the VAE encoder is not scale-equivariant~\cite{kouzelis2025eqvae}: $\mathcal{E}(\texttt{Resize}(\mathbf{x})) \neq \texttt{Resize}(\mathcal{E}(\mathbf{x}))$. The scale alignment must therefore be performed in image space, before encoding. We discuss these properties thoroughly in Section~\ref{sec:scale_mismatch}.

Building on these insights, we introduce \textbf{MACRO} (\textbf{M}ulti-plane \textbf{A}ttention for \textbf{C}loseup \textbf{R}ender \textbf{O}ptimization), a training-free method that resolves the scale mismatch. Our key observation is that diffusion-based enhancement models are 2D networks trained with 2D image priors, yet reference matching in close-up rendering is fundamentally a problem of geometric scale. While these 2D networks learn implicit 3D representations, their features are not scale-invariant, meaning they cannot reliably establish cross-view correspondences when the same content appears at drastically different zoom levels. 
Rather than forcing the network to overcome this mismatch purely in latent space, we propose MACRO, a training free light weight alternative. MACRO incorporates explicit 3D knowledge by leveraging the known 3D structure of the scene to decompose the rendered close-up into depth planes.
For each depth plane, we compute scale-matched crops of the reference images in image space before VAE encoding, and apply a depth-aware attention mask so that each close-up token attends only to reference tokens at the matching scale. This naturally supports multiple reference views, with each depth plane attending to the most informative reference. MACRO requires no architectural changes or training and operates as a preprocessing and attention masking strategy on top of existing enhancement models.

We demonstrate MACRO on top of the Difix~\cite{wu2025difix3d} architecture and evaluate it on two close-up benchmarks we construct for this task. 
As shown in Figure~\ref{fig:teaser}, MACRO produces faithful close-up renderings where baselines fail. Quantitatively, MACRO improves perceptual metrics by up to $38\%$ over the best enhancement baseline and by up to $67\%$ over unenhanced 3DGS, while maintaining competitive reconstruction scores.
To summarize, we make the following contributions:
\begin{itemize}
  \item We analyze the sensitivity of diffusion-based enhancement to scale, revealing two key properties: (a) VAE and U-Net features are not scale-invariant, causing cross-view attention to fail under scale mismatch, and (b) the VAE encoder is not scale-equivariant, preventing correction in latent space and necessitating image-space solutions.
  \item We propose MACRO, a training-free, depth-based multi-plane attention mechanism that resolves scale mismatch by cropping and resizing reference images in image space before encoding, allowing correct cross-view attention.
  \item We introduce two close-up novel view synthesis benchmarks, \textbf{DL3DV-Closeup}, curated from DL3DV-10K~\cite{ling2024dl3dv}, and \textbf{MobileClose-10}, captured specifically for this task, providing the first standardized evaluation protocol for close-up rendering under significant scale changes.
  \item We achieve state-of-the-art close-up rendering quality across both benchmarks, outperforming both 3DGS baselines and other diffusion-based approaches on reconstruction and perceptual metrics, while requiring no additional training.
\end{itemize}

\section{Related work}
\label{sec:related}

\textbf{High-fidelity novel view synthesis.}
3D Gaussian Splatting (3DGS)~\cite{kerbl3Dgaussians} enables real-time novel view synthesis, but rendering quality degrades for viewpoints far from the training cameras. Several works improve rendering fidelity without generative priors: Mip-Splatting~\cite{Yu2024MipSplatting} and multi-scale Gaussians~\cite{yan2024multiscale} address aliasing across zoom levels, FreGS~\cite{zhang2024fregs} improves high-frequency detail through frequency regularization, and Scaffold-GS~\cite{lu2024scaffoldgs} and Analytic Splatting~\cite{liang2024analytic} improve rendering precision. However, these methods cannot recover detail never observed in the training views, a fundamental limitation for close-up novel views captured far from any training camera.

\textbf{Close-up novel view synthesis.}
Recent work directly targets the close-up rendering problem. \cite{xia2025pseudolabel} generates pseudo-label supervision for close-up viewpoints and retrains the 3DGS representation. Close-up-GS~\cite{xia2025closeupgs} progressively self-trains to refine Gaussians for close-up views. CloseUpShot~\cite{zhang2026closeupshot} trains a point-conditioned diffusion model for close-up synthesis from sparse inputs. These methods all require retraining the reconstruction or a new generative model, which is both resource demanding and slow, while side-stepping the fundamental issues MACRO targets.

\textbf{Reference-guided rendering enhancement.}
Recent diffusion-based methods condition on reference images from the training set through attention layers to improve novel view quality. One line of work enhances 3DGS renderings by transferring textures from clean references~\cite{wu2023reconfusion,wu2025difix3d,fischer2025flowr,delutio2026artifixer,yin2025gsfixer}. A second group takes a purely generative approach, synthesizing novel views without relying on a 3DGS reconstruction~\cite{zhou2025seva,yu2024viewcrafter,gao2024cat3d}. For both groups, close-up views present an out-of-distribution case: the attention between the target view and references operates under a scale mismatch, and purely generative approaches must additionally generalize to camera poses unseen during training. \methodname{} addresses this scale mismatch directly.

\textbf{Scale equivariance in neural networks.}
Most vision model architectures are not scale-equivariant
~\cite{sosnovik2021disco,kouzelis2025eqvae}.
Existing remedies include scale-equivariant architectures
~\cite{worrall2019dss,sosnovik2021disco,rahman2023truly}
and equivariance regularization~\cite{kouzelis2025eqvae},
but these have not been demonstrated at the scale of the pretrained VAE/U-Net stacks used by reference-conditioned enhancement, and applying them would require retraining the enhancement backbone. \methodname{} instead resolves the scale mismatch in image space before encoding, requiring no retraining.

\textbf{Super-resolution for 3D Gaussian Splatting.}
A related line of work renders 3DGS at resolutions higher than the training images~\cite{feng2024srgs,xie2024supergs,asthana2025splatsure}, using sub-pixel constraints, latent feature fields, or 2D SR priors for multi-view consistent upsampling. These methods target uniform resolution upscaling from the same viewpoints. In contrast, close-up novel views create spatially-varying scale changes (foreground magnifies more than background) and reveal previously occluded content, motivating our multi-plane decomposition.

\section{Preliminaries: Reference-guided Enhancement}
\label{sec:preliminaries}

Several recent methods~\cite{wu2025difix3d,fischer2025flowr,delutio2026artifixer,yin2025gsfixer} enhance novel view renderings from 3DGS by conditioning a diffusion model on real reference images from the training set. In all of these methods, the rendered target view and reference image are encoded into a shared latent space, and an attention mechanism fuses information between the two. Since we build our method on top of Difix~\cite{wu2025difix3d}, we describe its specific formulation below and present empirical results on its SD-Turbo U-Net backbone.

Given a target view $\tilde{\mathbf{I}}_\text{target}$ rendered from the 3DGS reconstruction and a clean reference image $\mathbf{I}_{\text{ref}}$ captured by a training camera, both are encoded into latent representations by a VAE encoder $\mathcal{E}$:
\begin{equation}
\mathbf{z}_\text{target} = \mathcal{E}(\tilde{\mathbf{I}}_\text{target}), \quad \mathbf{z}_{\text{ref}} = \mathcal{E}(\mathbf{I}_{\text{ref}}).
\end{equation}
The latent tokens are processed by the U-Net with \emph{reference mixing layers}. In each such layer, the target and reference tokens are concatenated along the token dimension and joint self-attention is applied, allowing every target token to attend to every reference token:
\begin{align}
[\mathbf{z}_\text{target}';\, \mathbf{z}_{\text{ref}}'] &= \texttt{SelfAttn}([\mathbf{z}_\text{target};\, \mathbf{z}_{\text{ref}}]), \label{eq:selfattn}
\end{align}
where $[\cdot\,;\,\cdot]$ denotes concatenation along the token dimension. This enables the transfer of textures and details from the clean reference to the degraded rendering. Finally, the final predicted noise is used to analytically compute the cleaned latent $\mathbf{z'}_\text{cleaned}$, which is passed through the VAE's decoder to produce the final enhanced output $\hat{\mathbf{I}}_\text{target} = \mathcal{D}(\mathbf{z'}_\text{cleaned})$.

\section{Method}
\label{sec:method}
\begin{figure}[t]
  \centering
  \includegraphics[width=\textwidth]{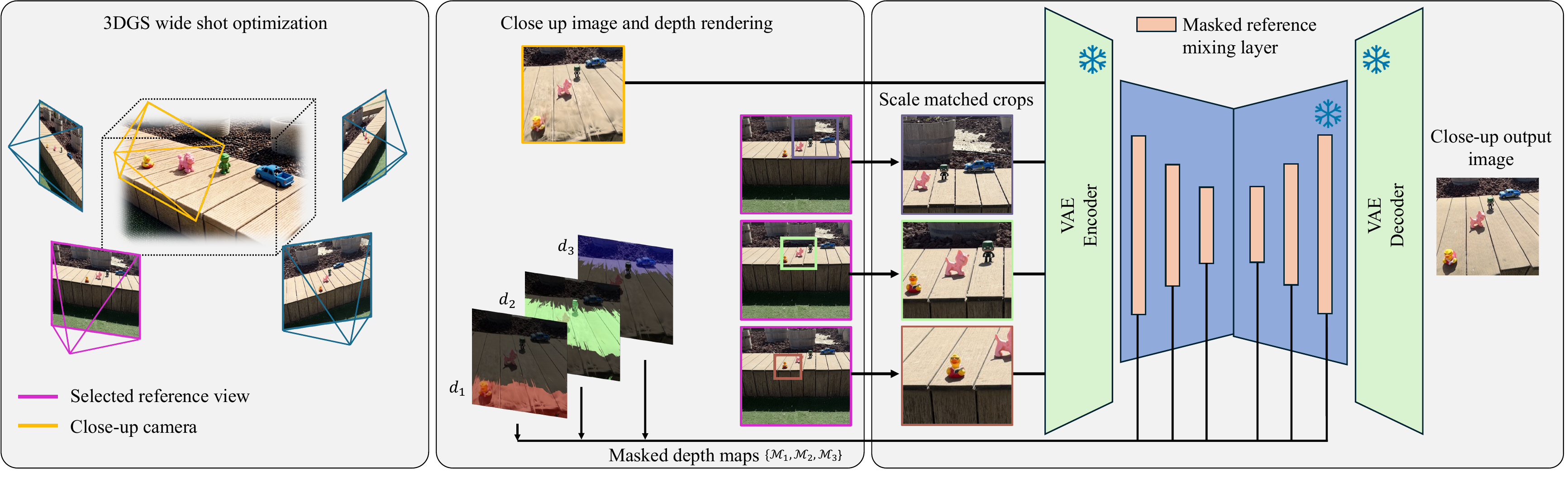}
  \caption{\textbf{\methodname{} pipeline.} Given a close-up rendering (\textbf{\textcolor{orange}{orange}}) and its depth map from the 3DGS reconstruction, we decompose the scene into $P$ depth planes with centroid depths $d_1, d_2, \ldots, d_P$, yielding corresponding depth masks $\mathcal{M}_1, \mathcal{M}_2, \ldots, \mathcal{M}_P$. For each depth plane, we compute a scale-matched crop of the reference image (\textbf{\textcolor{magenta}{pink}}) via depth-guided cropping and resizing, producing $P$ reference crops per reference view (color coded according to the depth planes). The close-up input and scale-matched reference crops are encoded by the VAE and processed through a masked reference mixing layer, where each close-up token attends only to reference tokens from the corresponding depth plane. The VAE decoder produces the enhanced close-up output. The figure illustrates the single-reference case; when using $K$ references, this yields $K \times P$ crops (Section~\ref{sec:reference_cropping}).}
  \label{fig:pipeline}
\end{figure}

Given a 3DGS reconstruction of a scene, our goal is to enhance a close-up view $\tilde{\mathbf{I}}_\text{cu}$ rendered from a close-up camera $\pi_\text{cu}$, using $N$ reference images $\{\mathbf{I}_{\text{ref}}^n\}_{n=1}^N$ captured by training cameras $\{\pi_{\text{ref}}^n\}_{n=1}^N$ and the depth map $\mathbf{D}_\text{cu}$ rendered from the 3DGS reconstruction. Figure~\ref{fig:pipeline} provides an overview of the \methodname{} pipeline. In Section~\ref{sec:scale_mismatch}, we analyze the scale mismatch problem that motivates our approach. We then present the three components of \methodname{}: multi-plane depth decomposition (Section~\ref{sec:depth_decomposition}), scale-matched reference cropping (Section~\ref{sec:reference_cropping}), and depth-masked attention (Section~\ref{sec:attention_masking}).

\subsection{Motivation: Scale mismatch in close-up rendering}
\label{sec:scale_mismatch}

We first analyze why existing reference-guided enhancement fails in the close-up setting. We identify two properties: (1) the learned features are sensitive to scale, causing attention to retrieve incorrect correspondences, and (2) this mismatch cannot be corrected in latent space due to the non-equivariance of the VAE encoder.

The reference mixing mechanism of Eq.~\eqref{eq:selfattn} implicitly assumes that corresponding content in the target and reference views appears at a similar spatial scale: tokens representing the same scene content should produce similar features so that attention retrieves the correct match. This assumption holds when both views have comparable fields of view. However, when the camera moves closer to the scene, objects appear at a much larger scale than in the reference images, and the same content is encoded into different feature representations, breaking the attention matching.

\begin{figure}[t]
  \centering
  \includegraphics[width=\textwidth]{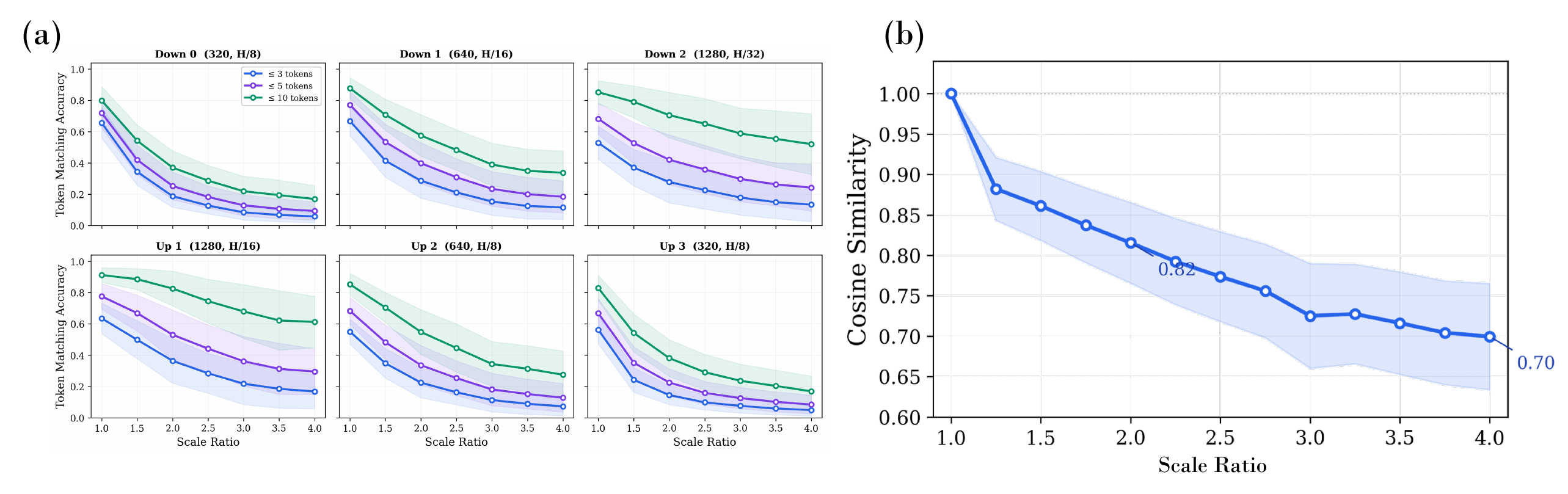}
  \caption{\textbf{Scale sensitivity analysis.} \textbf{(a)} Fraction of U-Net attention tokens whose nearest-neighbor key retrieval falls within a radius of 3, 5, or 10 tokens of the ground-truth correspondence, across six attention layers (Down 0--2, Mid, Up 1--2). Matching accuracy degrades at all layers as scale increases, confirming that cross-view attention fails under scale mismatch. Note that at coarser layers (Down2, Up1) the token matching accuracy is higher as there are less tokens overall. \textbf{(b)} Cosine similarity between the VAE latent of a resized image and the resized VAE latent of the original, i.e., $\mathcal{E}(\texttt{Resize}_s(\mathbf{x}))$ vs.\ $\texttt{Resize}_s(\mathcal{E}(\mathbf{x}))$. Similarity drops from 1.0 to 0.70 at 4$\times$ scale, showing that the VAE is not scale-equivariant and scale correction in latent space is less effective.}
  \label{fig:scale_analysis}
\end{figure}

\begin{figure}[t]
  \centering
  \includegraphics[width=0.495\textwidth, trim=0 149 0 0, clip]{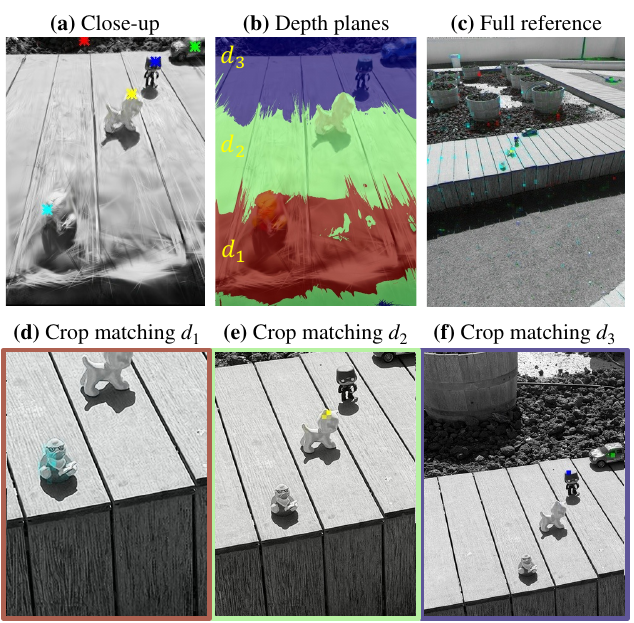}\hfill
  \includegraphics[width=0.495\textwidth, trim=0 0 0 150, clip]{figures/attn_labeled.pdf}
  \caption{\textbf{Attention mask visualization.} \textbf{(a)}~A rendered close-up with color-coded query tokens. \textbf{(b)}~Depth plane assignments. \textbf{(c)}~Standard (unmasked) attention maps from each query token to the full reference image; the duck query token incorrectly attends to the wood deck, explaining the texture artifacts visible in Difix results (Figure~\ref{fig:teaser}). \textbf{(d--f)}~Our masked attention constrains each query token to attend only to tokens on the scale-matched crop of its depth plane. Within each crop, the attention correctly retrieves the corresponding scene content.}
  \label{fig:attn_mask}
\end{figure}

\paragraph{Attention fails under scale mismatch.}
To quantify this effect, we simulate the close-up/reference scenario by cropping a sub-region from a native 4K image and resizing it to the processing resolution (emulating a close-up), while separately resizing the full image to the same resolution (emulating a reference). We encode both views with the VAE, run each through the U-Net, and at each self-attention layer compute the query projection $\mathbf{Q} = W_q(\mathbf{h})$ from the close-up and the key projection $\mathbf{K} = W_k(\mathbf{h})$ from the reference. For each close-up query token, we retrieve its nearest neighbor in the reference key space and measure the spatial displacement from the ground-truth correspondence. Figure~\ref{fig:scale_analysis}(a) reports the fraction of tokens whose match falls within 3, 5, or 10 tokens of the correct position across six U-Net attention layers. Matching accuracy degrades monotonically with increasing scale ratio, confirming that the pretrained attention cannot establish reliable correspondences under scale mismatch. We visualize this in Figure~\ref{fig:attn_mask}, showing how attention to the full reference retrieves incorrect correspondences, while our scale-matched crops restore correct matching.

\paragraph{Scale correction must happen in image space.}
One might attempt to fix this mismatch through geometric constraints on the attention, e.g., by warping or rescaling the latent representations after encoding. However, this approach is incorrect because the VAE encoder $\mathcal{E}$ is not scale-equivariant~\cite{kouzelis2025eqvae}: encoding a resized image does not produce the same result as resizing the encoded latent:
\begin{equation}
\label{eq:not_equivariant}
\mathcal{E}(\texttt{Resize}_s(\mathbf{x})) \neq \texttt{Resize}_s(\mathcal{E}(\mathbf{x})),
\end{equation}
where $\texttt{Resize}_s$ denotes spatial rescaling by factor $s$. We verify this in Figure~\ref{fig:scale_analysis}(b), which measures the cosine similarity between $\mathcal{E}(\texttt{Resize}_s(\mathbf{x}))$ and $\texttt{Resize}_s(\mathcal{E}(\mathbf{x}))$ as a function of scale factor $s$ on 200 images from DL3DV-10K~\cite{ling2024dl3dv}. This leads to the core principle underlying \methodname{}: scale alignment must be performed in image space, before VAE encoding.

\subsection{Multi-plane depth decomposition}
\label{sec:depth_decomposition}

Given the analysis above, we now present the design of \methodname{}. In a close-up view, objects at different depths appear at different scales relative to the reference images. A foreground object close to the camera undergoes a larger scale change than a background object. To handle this, we decompose the close-up rendering into $P$ depth planes, where each plane groups pixels at similar depths and thus similar scale factors.
Given the rendered depth map $\mathbf{D}_\text{cu}$, we partition the pixels into $P$ groups using $k$-means clustering on the depth values:
\begin{equation}
\{d_1, d_2, \ldots, d_P\} = \texttt{k\text{-}means}(\mathbf{D}_\text{cu}, P),
\end{equation}
where $d_p$ is the centroid depth of plane $p$. Each pixel $i$ in the close-up view is assigned to its nearest depth plane, yielding a partition $\{\mathcal{M}_1, \mathcal{M}_2, \ldots, \mathcal{M}_P\}$ of the close-up image, where $\mathcal{M}_p$ is the set of pixel indices belonging to plane $p$. Clustering is performed in the reciprocal $1/z$-component space.

Given that $N$ views were used to train the 3DGS to reconstruct the full scene, most of them are not useful when enhancing a very specific close-up. We therefore apply per close-up a greedy choice of $\{\mathbf{I}_{\text{ref}}^k\}_{k=1}^K$  references that cover the input angle best where $K\le N$, and also eliminate references who do not see the close-up due to occlusion. Then only these $K$ references are used for the enhancement pipeline. We found that setting $K=3$ references and $P=3$ depth planes to be optimal and use these values throughout our experiments. More details about the hyperparameter search are in Section~\ref{sec:ablation}.

\subsection{Scale-matched reference cropping}
\label{sec:reference_cropping}

For each depth plane $p$ with centroid depth $d_p$, we compute a scale-matched crop of each reference image that shows the corresponding scene content at the same scale as it appears in the close-up view. Given $K$ reference views and $P$ depth planes, this produces a total of $K \times P$ reference crops.
\paragraph{Scale factor.} For depth plane $p$, the scale factor between the close-up and reference view is determined by the ratio of the object's depth in the reference to its depth in the close-up:
\begin{equation}
\label{eq:scale_factor}
s_p = \frac{d_p^{\text{ref}}}{d_p^{\text{cu}}},
\end{equation}
where $d_p^{\text{cu}}$ and $d_p^{\text{ref}}$ are the depths of the plane-$p$ content as seen from the close-up and reference cameras, respectively. Objects closer to the close-up camera (small $d_p^{\text{cu}}$) yield a larger scale factor, meaning they appear proportionally larger in the close-up than in the reference. The crop size in the reference image is set to cover a region of $1/s_p$ of the reference field of view, so that after resizing to the processing resolution, the content appears at the same scale as in the close-up.

\paragraph{Crop location and upsampling.} We project the 3D points at depth $d_p$ visible in the close-up into the reference camera and center the crop on the projected region. The crop is upsampled to match the input resolution (1k in our experiments) using a state of the art super-resolution model PFT-SR \cite{long2025pftsr} while maintaining aspect-ratio. The resulting scale-matched reference patch $\mathbf{I}_{\text{ref}}^{k,p}$ is encoded as $\mathbf{z}_{\text{ref}}^{k,p} = \mathcal{E}(\mathbf{I}_{\text{ref}}^{k,p})$. Because cropping and resizing occur in image space before encoding, the attention mechanism can establish correct correspondences (Section~\ref{sec:scale_mismatch}).

\subsection{Depth-masked Multi-plane Attention}
\label{sec:attention_masking}

The close-up render and all $K \times P$ scale-matched reference crops are encoded through the VAE and processed jointly by the U-Net mixing blocks (Section~\ref{sec:preliminaries}).
To ensure correct scale texture copying, for each depth plane $p$ we construct an attention mask $\mathbf{A} \in \{0, 1\}^{S_l \times KS_l}$ that controls which close-up tokens can attend to the reference crops of that plane's scale, where $S_l=H_l\times W_l$ is the number of spatial latent tokens of a single view at mixing block $l$. Each close-up token from depth plane $p$ attends only to reference tokens from the corresponding scale-matched patch for plane $p$. Let $\mathcal{M}^p_l$ denote the set of close-up token indices belonging to plane $p$, and let $\mathcal{R}^{k,p}_l$ denote the reference token indices corresponding to the scale-matched crop for plane $p$ from reference $k$, both downsampled to the resolution of $H_l,W_l$. Then:
\begin{equation}
\label{eq:attn_mask}
\mathbf{A}[i, (k-1)S_l+j] = 1 \iff i \in \mathcal{M}^p_l \text{ and } j \in \mathcal{R}^{k,p}_l.
\end{equation}

We then construct the full multi-view attention mask $\mathbf{A'} \in \{0, 1\}^{(KP+1)S_l \times (KP+1)S_l}$, where per-view self attention is fully allowed for both input and reference views, cross-reference attention is blocked, and input-reference cross-view attention is determined by $\mathbf{A}$. The modified reference mixing layer (Eq.~\eqref{eq:selfattn}) then becomes:
\begin{equation}
\mathbf{z}'_l = \texttt{MaskedSelfAttn}(\mathbf{z}_l, \mathbf{A'}),
\end{equation}
where $\mathbf{z}_l = [\mathbf{z}_\text{cu};\, \mathbf{z}_{\text{ref}}^{1,p};\, \ldots;\, \mathbf{z}_{\text{ref}}^{K,P}]$ is the concatenation of the close-up and all reference latents at U-Net mixing block $l$, and the mask $\mathbf{A}$ is applied to the attention scores before the softmax. Notably, this requires only a single forward pass through the diffusion model. The mask controls the information flow without requiring multiple runs.

\section{Experiments}
\label{sec:experiments}

We evaluate \methodname{} on two close-up novel view synthesis benchmarks that we construct as part of this work, addressing the lack of standardized evaluation protocols for this setting. In Section~\ref{sec:benchmarks_metrics}, we describe the evaluation benchmarks and metrics. In Appendix~\ref{app:implementation}, we provide implementation details. In Section~\ref{sec:main_results}, we compare \methodname{} against existing methods. In Section~\ref{sec:ablation}, we ablate key design choices. In Section~\ref{sec:limitations}, we discuss limitations.

\subsection{Benchmarks and Metrics}
\label{sec:benchmarks_metrics}

\textbf{Benchmarks.}
Existing novel view synthesis benchmarks do not systematically evaluate close-up rendering, where the target camera is significantly closer to the scene than any training view. To address this gap, we introduce two new benchmarks that test different aspects of close-up novel view synthesis.
\textbf{DL3DV-Closeup} is our primary benchmark, comprising 44 diverse in-the-wild scenes curated from DL3DV-10K~\cite{ling2024dl3dv} by selecting scenes with large depth disparity between camera positions, yielding naturally occurring close-up/far view pairs with a total of 314 evaluation pairs.
Our newly curated \textbf{MobileClose-10} contains 10 scenes captured specifically for the close-up task using a smartphone camera, and serves to evaluate generalization beyond DL3DV with 39 far/close evaluation pairs.
Full details on the construction of both benchmarks are provided in Appendix~\ref{sec:benchmark_details}.

\textbf{Metrics.}
We report PSNR and SSIM as standard pixel-aligned reconstruction metrics. However, these measures do not fully capture whether the enhanced close-up faithfully reproduces the semantic content and fine-grained structure of the scene: two images can have similar PSNR yet differ substantially in perceptual fidelity, particularly when diffusion-based enhancement introduces plausible but spatially shifted details. We therefore additionally report three perceptual metrics: LPIPS~\cite{zhang2018perceptual}, DreamSim~\cite{fu2023dreamsim}, and DINOv2~\cite{oquab2023dinov2} feature distance. These operate in learned feature spaces and better reflect whether the output preserves the identity and structure of the original scene content.

\textbf{Baselines.}
We compare against the reconstruction methods 3DGS~\cite{kerbl3Dgaussians} and Mip-Splatting~\cite{Yu2024MipSplatting}, the purely generative NVS method SEVA~\cite{zhou2025seva}, and the
   enhancement methods GSFixer~\cite{yin2025gsfixer} and Difix~\cite{wu2025difix3d}. We also evaluate progressive variants that attempt to bridge the distribution gap of the close-up setting gradually, as
   discussed in Section~\ref{sec:intro}: Difix3D progressively renders and enhances intermediate views at incrementally closer distances, distilling results back into the 3DGS representation at each step.
   PR-Difix3D (Progressive Reference) also updates the reference images at each step to decrease the scale gap between the rendered and reference, implementing the strategy of
   Close-up-GS~\cite{xia2025closeupgs} which did not release code. The "+" variants (Difix3D+, PR-Difix3D+) apply an additional enhancement step at the final iteration. Our GSFixer setup similarly reflects the interpolated trajectory strategy of CloseUpShot~\cite{zhang2026closeupshot} (see Appendix~\ref{app:baseline_details}), as its implementation was also unavailable at the time of submission. FlowR~\cite{fischer2025flowr} and ArtiFixer~\cite{delutio2026artifixer} were not available either.

\subsection{Results}
\label{sec:main_results}

\begin{figure}[t]
  \centering
  \includegraphics[width=\textwidth]{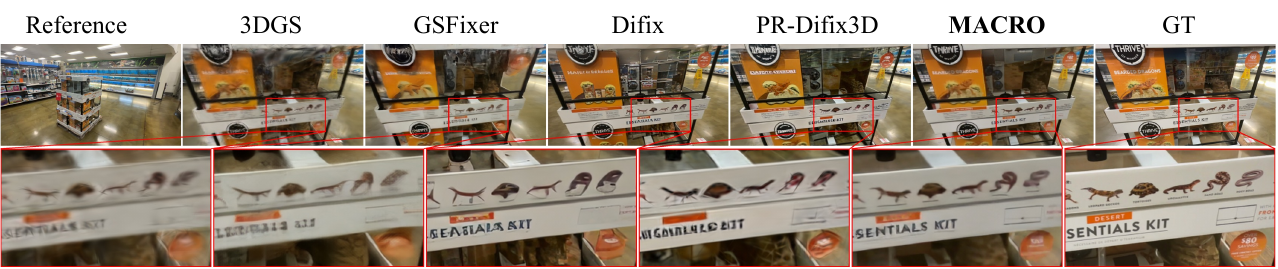}
  \caption{\textbf{Qualitative results on DL3DV-Closeup.} Only in \methodname{} is the text on the "Essentials Kit" readable and the animal figures faithfully preserved (see crops).}
  \label{fig:results_dl3dv}
\end{figure}

\begin{figure}[t]
  \centering
  \includegraphics[width=\textwidth]{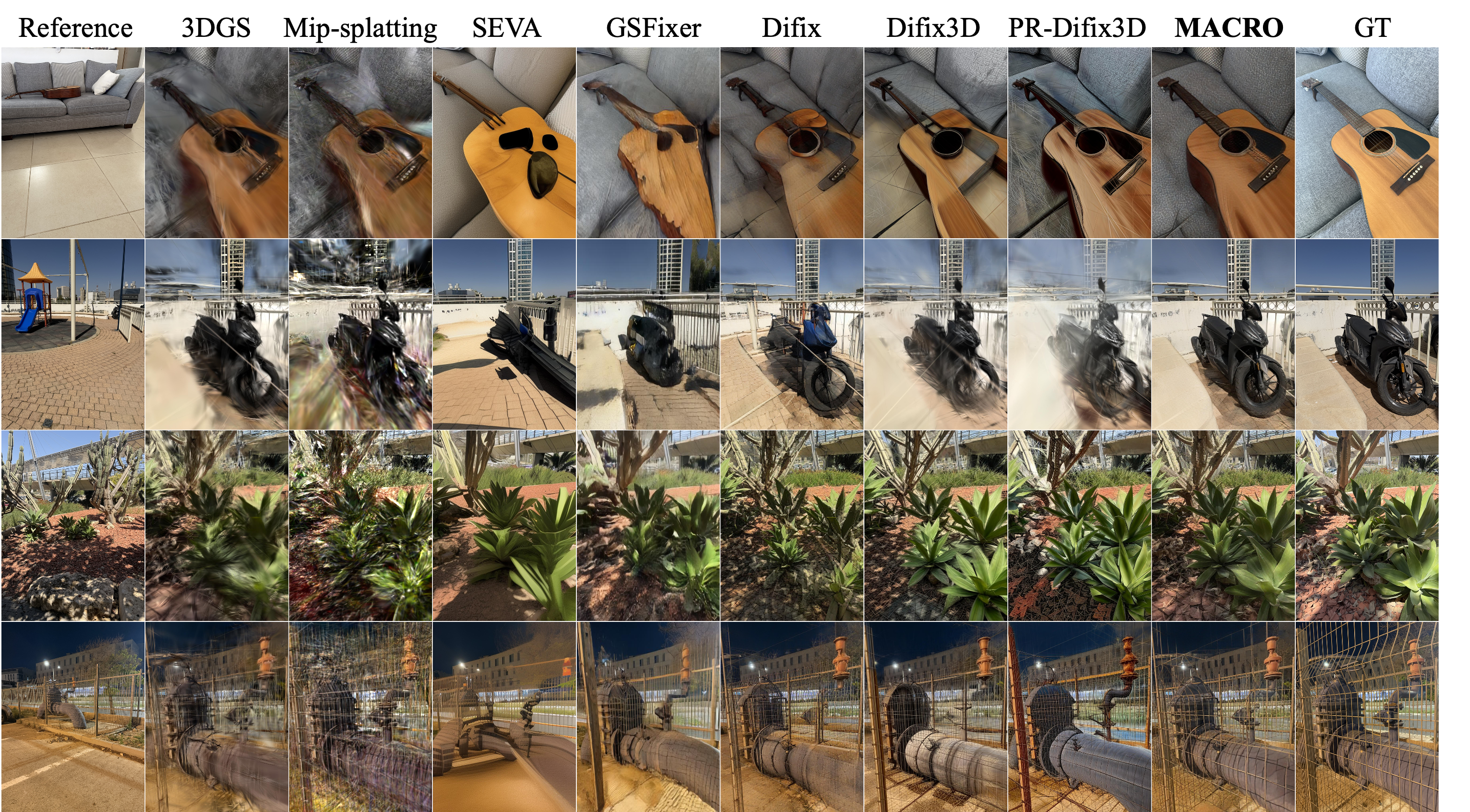}
  \caption{\textbf{Qualitative results on MobileClose-10.} \methodname{} produces close-up renderings that faithfully match the ground truth (up to exposure differences, as described in Appendix~\ref{sec:benchmark_details}), while baseline methods either remain blurry or transfer incorrect textures from the reference images. Best viewed zoomed-in on a digital screen.}
  \label{fig:results}
\end{figure}

Table~\ref{tab:main_results} reports quantitative results on both benchmarks and Figures~\ref{fig:results_dl3dv} and~\ref{fig:results} show qualitative comparisons on DL3DV-Closeup and MobileClose-10 respectively (additional qualitative results on DL3DV-Closeup are provided in Appendix~\ref{app:qualitative}). Although 3DGS and Mip-Splatting avoid the texture artifacts of enhancement methods, their renderings remain blurry and lack fine detail at close range. SEVA produces sharp outputs but diverges from the ground truth, likely because close-up views are out of distribution relative to its training set and it relies solely on camera parameters without leveraging the scene's 3DGS reconstruction. Difix improves perceptual metrics over 3DGS, confirming that the diffusion model adds detail, but it copies wrong textures from the reference, and due to the scale difference, even when the attention mechanism looks at the correct location, the textures are transferred at the wrong scale (see Figure~\ref{fig:results} last two rows - cactus and fence scales). Difix3D does not improve, as the progressive enhancement eventually reaches the wrong-scale regime and the inconsistent intermediate views degrade the 3DGS model. PR-Difix3D updates the reference progressively but does not consistently improve over Difix3D, and remains far below \methodname{} on all perceptual metrics.

\methodname{} achieves the best perceptual metrics on both benchmarks while maintaining competitive PSNR, confirming that scale-matched references enable the model to transfer correct scene content rather than hallucinating plausible but misaligned details. Results are consistent across DL3DV-Closeup and MobileClose-10, demonstrating generalization.

\begin{table}[t]
\centering
\caption{\textbf{Quantitative comparison on both benchmarks.} DSim = DreamSim, DINO = DINOv2 feature distance. \methodname{} achieves the best perceptual metrics on both benchmarks while maintaining competitive reconstruction scores.}
\label{tab:main_results}
\resizebox{\textwidth}{!}{%
\begin{tabular}{l@{\hspace{6pt}}ccccc@{\hspace{4pt}}|@{\hspace{4pt}}ccccc}
\toprule
& \multicolumn{5}{c}{\textit{DL3DV-Closeup (44 scenes)}} & \multicolumn{5}{c}{\textit{MobileClose-10 (10 scenes)}} \\
\cmidrule(r){2-6} \cmidrule(l){7-11}
Method & PSNR$\uparrow$ & SSIM$\uparrow$ & LPIPS$\downarrow$ & DSim$\downarrow$ & DINO$\downarrow$ & PSNR$\uparrow$ & SSIM$\uparrow$ & LPIPS$\downarrow$ & DSim$\downarrow$ & DINO$\downarrow$ \\
\midrule
3DGS \cite{kerbl3Dgaussians}         & 16.00 & 0.499 & 0.457 & 0.261 & 0.462 & \underline{14.58} & \textbf{0.441} & 0.578 & 0.293 & 0.512 \\
Mip-Splatting \cite{Yu2024MipSplatting} & \textbf{16.26} & \textbf{0.514} & 0.478 & 0.257 & 0.440 & 14.41 & 0.428 & 0.667 & 0.335 & 0.521 \\
SEVA \cite{zhou2025seva}          & 12.35 & 0.335 & 0.544 & 0.156 & 0.263 & 13.91 & 0.401 & 0.647 & 0.270 & 0.417 \\
GSFixer  \cite{yin2025gsfixer}     & 15.91 & 0.494 & 0.378 & \underline{0.121} & 0.250 & 14.51 & 0.401 & 0.614 & \underline{0.250} & 0.408 \\
\midrule
Difix   \cite{wu2025difix3d}      & 15.81 & 0.477 & \underline{0.343} & 0.128 & \underline{0.205} & 14.55 & 0.407 & \underline{0.557} & 0.253 & \underline{0.352} \\
Difix3D    \cite{wu2025difix3d}      & 14.69 & 0.427 & 0.394 & 0.149 & 0.248 & 13.76 & 0.418 & 0.582 & 0.258 & 0.406 \\
Difix3D+   \cite{wu2025difix3d}      & 14.33 & 0.404 & 0.393 & 0.155 & 0.243 & 13.36 & 0.385 & 0.579 & 0.270 & 0.394 \\
PR-Difix3D    & 15.15 & 0.433 & 0.399 & 0.153 & 0.262 & 13.68 & 0.393 & 0.563 & 0.292 & 0.428 \\
PR-Difix3D+   & 14.42 & 0.379 & 0.430 & 0.178 & 0.271 & 13.00 & 0.329 & 0.596 & 0.320 & 0.437 \\
\midrule
\textbf{MACRO (ours)} & \underline{16.17} & \underline{0.505} & \textbf{0.330} & \textbf{0.105} & \textbf{0.152} & \textbf{14.60} & \underline{0.433} & \textbf{0.551} & \textbf{0.168} & \textbf{0.217} \\
\bottomrule
\end{tabular}}
\end{table}

\subsection{Ablation studies}
\label{sec:ablation}

We ablate key design choices of \methodname{} on DL3DV-Closeup (Table~\ref{tab:ablation}). Reducing from $K{=}3$ to $K{=}1$ references drops perceptual metrics, confirming that multiple references provide complementary coverage. To isolate the contribution of scale matching from the multi-plane mechanism, we evaluate a \emph{single-plane} variant ($P{=}1$, no mask) that applies one global scale-matched crop per reference. This variant alone already outperforms all enhancement baselines on every perceptual metric (Table~\ref{tab:main_results}), confirming that image-space scale correction is the primary driver. The full multi-plane formulation improves further, and its gain grows with the depth range within the close-up: on close-ups with a near-to-far depth ratio of at least 10, \methodname{} improves DreamSim by $10.5\%$ and DINOv2 by $10.4\%$ over the single plane (Appendix~\ref{app:single_plane}), since a single crop can match only one scale. Notably, adding planes \emph{without} the mask performs worse than a single crop, showing that the mask is what lets each token use the reference at its own scale. Replacing PFT-SR with LANCZOS upsampling yields comparable PSNR but worse perceptual scores, and backward-warping the reference instead of cropping causes large degradation, as depth noise in the warp destroys high-frequency texture. Finally, \methodname{} is robust to depth noise: perturbing the rendered depth by up to $20\%$ changes LPIPS by at most $0.007$ and DreamSim and DINOv2 by at most $0.001$ (Appendix~\ref{app:depth_robustness}).

\begin{figure}[t]
  \centering
  \includegraphics[clip,trim=0 1.5cm 0 1.3cm,width=\textwidth]{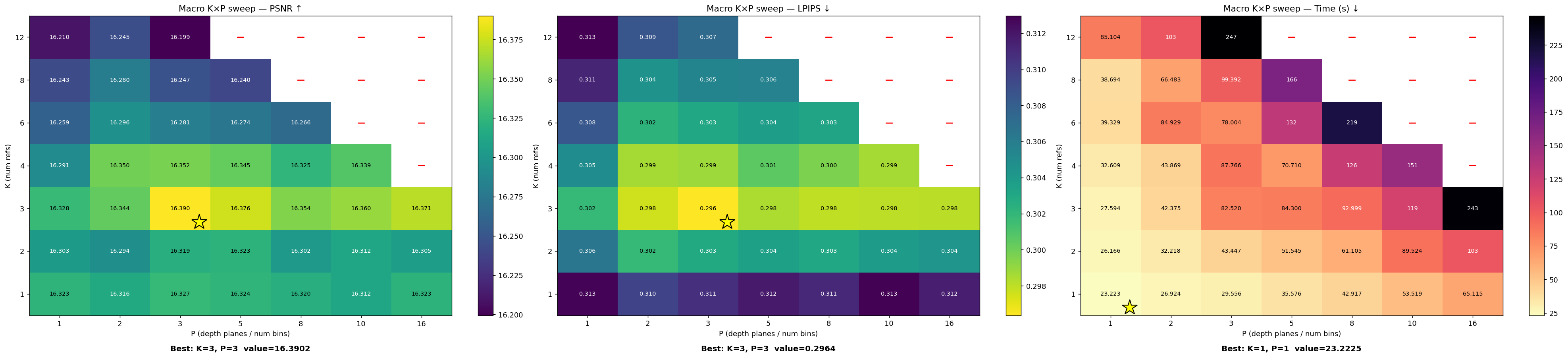}
  \caption{\textbf{Hyperparameter search of $K,P$}. \textbf{left}: PSNR, \textbf{middle}: LPIPS, \textbf{right}: end to end runtime in seconds. We observe optimal LPIPS,PSNR for $K=P=3$, while configurations with $KP\ge 60$ cause OOM errors.}
  \label{fig:heatmaps}
\end{figure}

To choose the number of references $K$ and number of depth planes $P$ we performed grid search over a representative subset of 8 scenes from DL3DV-Closeup, as shown in Figure~\ref{fig:heatmaps}. We evaluated $K\in\{1,2,3,4,6,8,12\}$ and
$P\in\{1,2,3,5,8,10,16\}$. We found that setting $K=P=3$ resulted in maximal PSNR and minimal LPIPS at a moderate runtime; increasing $P$ has diminishing returns on quality. For $KP\ge 60$ we encountered OOM issues which limited the search, and higher $KP$ values also increased runtime - mainly attributed to the enhancer itself, as attention is quadratic in nature. The rest of the pipeline (reference selection, warping, cropping and super resolution) are all relatively negligible in terms of processing time. We therefore use \methodname{} with $K=P=3$ throughout the paper.

\begin{table}[t]
\centering
\small
\caption{\textbf{Ablation study on DL3DV-Closeup.} Each row varies a single component from the full \methodname{} pipeline. $K,P$: number of reference views and depth planes; $P{=}1$ uses a single global scale-matched crop per reference. Mask: depth-aware attention masking. Upsample: method used to resize reference crops, SR = PFT-SR~\cite{long2025pftsr}.}
\label{tab:ablation}
\setlength{\tabcolsep}{4pt}
\begin{tabular}{lccc@{\hspace{8pt}}ccccc}
\toprule
& \multicolumn{3}{c}{Components} & \multicolumn{5}{c}{Metrics} \\
\cmidrule(r){2-4} \cmidrule(l){5-9}
Config & $K,P$ & Mask & Upsample & PSNR $\uparrow$ & SSIM $\uparrow$ & LPIPS $\downarrow$ & DreamSim $\downarrow$ & DINOv2 $\downarrow$ \\
\midrule
\textbf{MACRO (full)} & $3,3$ & \checkmark & SR & 16.17 & 0.505 & \textbf{0.330} & \textbf{0.105} & \textbf{0.152} \\
\midrule
$K{=}1$          & $1,3$ & \checkmark & SR      & \underline{16.18} & \underline{0.506} & 0.349 & 0.120 & 0.185 \\
$P{=}1$          & $3,1$ & ---        & SR      & 16.13 & 0.501 & \underline{0.332} & \underline{0.108} & \underline{0.157} \\
No mask          & $3,3$ & ---        & SR      & 16.12 & 0.501 & 0.333 & 0.110 & 0.168 \\
LANCZOS          & $3,3$ & \checkmark & LANCZOS & \textbf{16.20} & \textbf{0.509} & 0.360 & \underline{0.108} & 0.164 \\
Warp             & $3,3$ & \checkmark & ---     & 15.97 & 0.474 & 0.399 & 0.211 & 0.412 \\
\bottomrule
\end{tabular}
\end{table}

\subsection{Limitations}
\label{sec:limitations}

MACRO crops and resizes references to match the close-up scale, but the upsampled crops lose high-frequency details, introducing a blur-realism tradeoff. A dedicated super-resolution stage on the final
   output could alleviate this. Additionally, the number of depth planes $P$ is currently fixed rather than adapted to the depth complexity of each view, leaving room for a per-image adaptive scheme. Moreover, MACRO relies on the 3DGS-rendered depth map for plane decomposition and crop computation, which may be inaccurate in regions with insufficient training views. Off-the-shelf depth estimators did not improve over the 3DGS depth, as they yield sparse depth at close range. Rarely, faint seams appear at plane boundaries; letting boundary tokens attend to adjacent planes is a simple remedy. Finally, distilling \methodname{} outputs back into the 3DGS with standard schemes degrades quality (Appendix~\ref{app:distillation}); close-up-aware 3DGS densification is a possible solution.

\section{Conclusion}
\label{sec:conclusion}
We addressed close-up novel view synthesis from 3DGS, where reference-conditioned diffusion enhancement fails because the underlying features are not scale-invariant and the VAE is not scale-equivariant,
   ruling out latent-space correction. MACRO resolves this by decomposing the close-up into depth planes and providing scale-matched reference crops through a depth-aware attention mask, injecting 3D knowledge
   into a 2D pipeline without architectural changes or finetuning. On two close-up benchmarks, MACRO improves perceptual metrics over the strongest
   enhancement baseline by up to $26\%$ and $38\%$ respectively, and by up to $67\%$ and $58\%$ over unenhanced 3DGS. As a training-free method, MACRO is readily applicable to settings where close-up training data is unavailable, such as fixed camera arrays used for dynamic scenes. 
   Given the plug and play nature of MACRO, we expect that leveraging it on top of stronger generative backbones would result in further improvements, closing the gap in close-up novel view synthesis.

\begin{ack}
This work was funded by Amazon. The authors did not receive any third-party funding or support for this work.
All authors were employed by Amazon (Prime Video) while conducting this work. The authors declare no competing interests beyond the affiliations listed.
\end{ack}

\bibliographystyle{plainnat}
\bibliography{references}

\clearpage
\appendix


\begin{center}
{\LARGE \textbf{MACRO: Training-free Multi-plane Attention for Closeup Render Optimization}}\\[4pt]
{\large ------------}\\[4pt]
{\LARGE \textbf{Supplementary Material}}
\end{center}


The supplementary material provides additional qualitative results, extended ablation studies, implementation details, and further details on the construction of the close-up benchmarks.

\section*{Appendix Contents:}
\begin{enumerate}[label=\Alph*.]
    \item Additional Qualitative Results (Appendix \ref{app:qualitative})
    \item Implementation Details (Appendix \ref{app:implementation})    
    \item Additional Ablations (Appendix \ref{app:ablations})
    \item Benchmark Construction Details (Appendix \ref{sec:benchmark_details})
\end{enumerate}
\clearpage

\section{Additional qualitative results}
\label{app:qualitative}

\begin{figure*}[h]
  \centering
  \includegraphics[width=\textwidth]{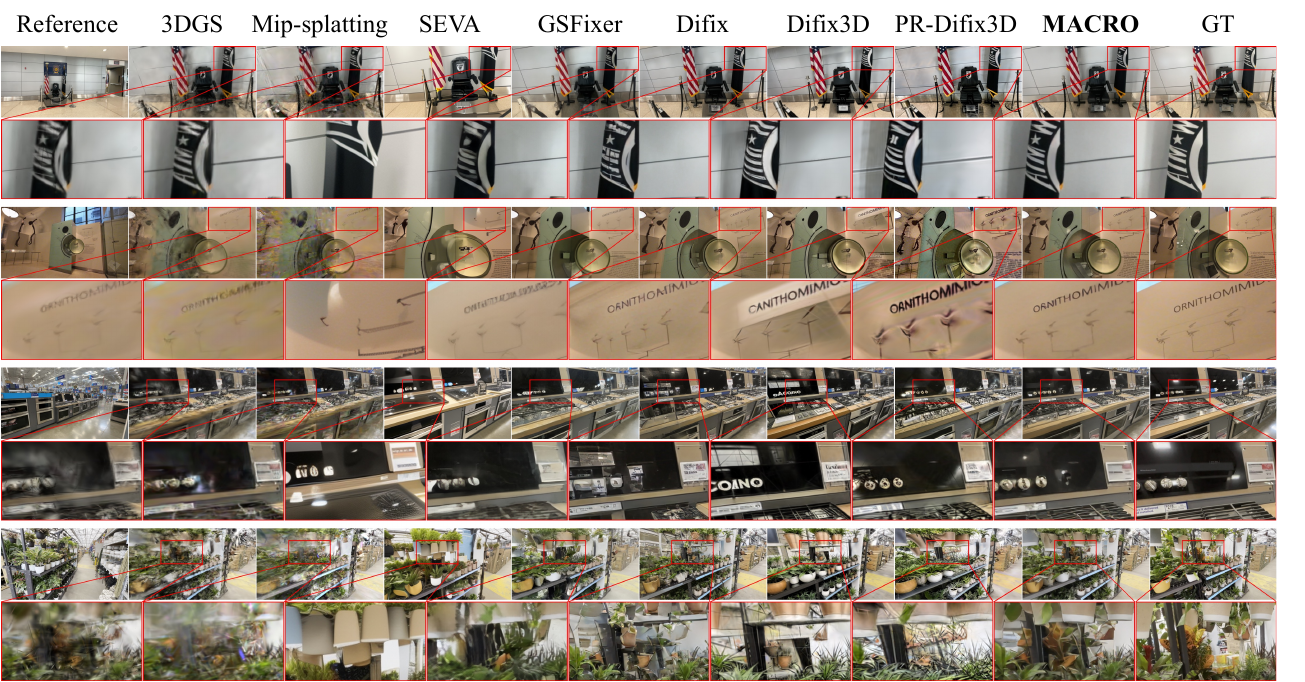}
  \caption{\textbf{Additional qualitative results on DL3DV-Closeup.} Four examples shown, each with a full-frame view and a crop below for detail. \methodname{} produces more faithful renderings that better match the ground truth. Best viewed zoomed-in on a digital screen.}
  \label{fig:results_disparity}
\end{figure*}

\section{Implementation details}
\label{app:implementation}
\subsection{MACRO implementation details}

\paragraph{Reference selection.}

For each close-up render, MACRO selects $K$ reference views from the training set by greedy set cover over pixel observability: each candidate's 3dgs rasterized depth map is unprojected to world coordinates and reprojected into the close-up camera, and references are added one at a time to maximize newly covered pixels. References who cannot see the close-up completely due to full occlusion determined by reference view depth are discarded. We use the 3DGS rasterized depth throughout; we tested alternative SOTA depth estimation methods for reference elimination module, including MVSAnywhere\cite{mvsanywhere} and Mast3r\cite{mast3r}, which produced negligible differences. These methods were not applicable to estimate input depth however, as the input close-up render is too corrupted.

\paragraph{Compute resources.}
Because MACRO is training-free, the total compute is dominated by inference: per close-up, the depth-plane construction, reference selection, cropping, and resizing add negligible overhead on top of the underlying enhancement backbone, and the super-resolution module adds a small fixed cost per crop. MACRO's per-view runtime is therefore close to that of its backbone~\cite{wu2025difix3d}, which is itself a single-step distilled diffusion model. All experiments were run on a single NVIDIA P5 instance over approximately one week. MACRO inference takes ${\sim}34$ seconds per image ($540\times 960$ resolution) on DL3DV-Closeup and ${\sim}104$ seconds on MobileClose-10 ($1071\times 1428$ resolution). These are much faster compared to the other solutions that either require multi-step video diffusion or iterative finetuning of the 3dgs representation, which can take over 30 minutes per scene.

\subsection{Baseline adaptations for close-up novel view synthesis}
\label{app:baseline_details}




\paragraph{Difix3D, Difix3D+, and Progressive-Reference variants.}
Unlike Difix \cite{wu2025difix3d}, which applies a single post-processing correction to the rendered close-up view at inference time, Difix3D \cite{wu2025difix3d} progressively refines the 3DGS representation by adding diffusion-corrected rendered views along the path that connects the training views to the close-up camera. In addition to the original implementation, we evaluate a variant that progressively updates the reference set, which we call \textbf{Progressive-Reference Difix3D (PR-Difix3D)}. After each refinement round, the newly corrected intermediate views are added as candidate references for subsequent iterations. \textbf{PR-Difix3D+}, is the analogous extension of Difix3D+.

\paragraph{GSFixer.}
GSFixer~\cite{yin2025gsfixer} is a video diffusion model based on a fine-tuned CogVideoX-5B~\cite{yang2025cogvideox}, designed to correct 3DGS rendering artifacts. To adapt GSFixer to our close-up benchmarks, we construct a 49-frame sequence for each target close-up view. Specifically, we select the two best training references using the same heuristic as Difix3D. We then interpolate frames from the first reference to the close-up view and from the close-up view to the second reference, placing the target close-up at the center of the sequence.

\section{Additional ablations}
\label{app:ablations}
\subsection{Single-plane ablation}
\label{app:single_plane}
To quantify how much of \methodname{}'s gain comes from scale correction alone, we evaluated a single-global-crop variant ($P{=}1$, $K{=}3$, no depth mask) against full \methodname{} on all 44 DL3DV-Closeup scenes (Table~\ref{tab:ablation}). Averaged over the benchmark, full \methodname{} improves over the $P{=}1$ variant by $+3.1\%$ on DreamSim ($0.108 \rightarrow 0.105$), $+3.0\%$ on DINOv2 ($0.157 \rightarrow 0.152$) and $+0.5\%$ on LPIPS ($0.332 \rightarrow 0.330$). This gap, however, is far from uniform: it concentrates exactly where multiple planes are needed. A single global crop can match only one scale, so it suffices for planar close-ups but fails where a view spans a wide range of depths. Stratifying by within-view depth disparity, the gain of full \methodname{} over the $P{=}1$ variant increases monotonically with the depth ratio. Averaged over the 33 close-ups in the highest stratum (depth ratio $\geq 10$), full \methodname{} improves DreamSim from $0.142$ to $0.127$ ($+10.5\%$), DINOv2 from $0.219$ to $0.196$ ($+10.4\%$), and LPIPS from $0.374$ to $0.365$ ($+2.3\%$). On individual views the effect is even more pronounced; for example, on a scene with depth ratio ${\approx}13$, DreamSim improves from $0.254$ to $0.180$ ($+29\%$), DINOv2 from $0.592$ to $0.448$ ($+24\%$), and LPIPS from $0.414$ to $0.390$ ($+6\%$).

We see this as reflecting two distinct contributions. First, the diagnostic analysis of Section~\ref{sec:scale_mismatch}: applying it as a single global crop, with no additional machinery, training, or 3DGS re-optimization, already outperforms the enhancement baselines (Table~\ref{tab:main_results}) on all three perceptual metrics by $+3.4\%$ to $+23.5\%$. Second, the full multi-plane formulation, which is the general form of scale matching rather than an add-on to it: because a close-up observes content across a range of depths, the correct reference scale is not global, and the single global crop is its degenerate planar special case ($P{=}1$). The median within-view depth ratio in DL3DV-Closeup is only $3.5\times$, so planar cases dominate the aggregate, but the full formulation is what keeps \methodname{} correct on the harder, non-planar cases where a single crop fails.

\subsection{Robustness to depth errors}
\label{app:depth_robustness}
\methodname{} relies on the 3DGS-rendered close-up depth, which can be inaccurate in under-observed regions. To quantify sensitivity, we perturb the rendered depth with multiplicative Gaussian noise, $\tilde{\mathbf{D}} = \mathbf{D}_\text{cu}(1+\sigma\mathcal{N}(0,1))$, and rerun the full pipeline (Table~\ref{tab:depth_noise}). Even at $\sigma{=}20\%$, metrics remain essentially unchanged, since depth enters only through aggregate statistics: per-plane median depths for the scale factors (Eq.~\eqref{eq:scale_factor}) and $k$-means centroids, pooled to latent-token resolution, for the mask (Eq.~\eqref{eq:attn_mask}). \methodname{} also performs a single feed-forward pass, so depth errors cannot accumulate across steps.

\textbf{Alternative depth sources.} We also tested off-the-shelf estimators, MVSAnywhere~\cite{mvsanywhere} and MASt3R~\cite{mast3r}, in place of the 3DGS depth. They gave equal or worse results: when fit to wide-shot images and queried at a close-up, their depth is sparse, and naive completion does not densify it correctly. Because the depth is used to partition the close-up render itself, the 3DGS depth is more consistent with that image. Close-up-specific depth completion could close this gap.

\begin{table}[h]
\centering
\small
\caption{\textbf{Depth perturbation on DL3DV-Closeup.} The rendered close-up depth is perturbed with multiplicative Gaussian noise of standard deviation $\sigma$.}
\label{tab:depth_noise}
\begin{tabular}{lccccccc}
\toprule
Config & Mask & $K,P$ & PSNR $\uparrow$ & SSIM $\uparrow$ & LPIPS $\downarrow$ & DreamSim $\downarrow$ & DINOv2 $\downarrow$ \\
\midrule
\methodname{} & \checkmark & $3,3$ & 16.17 & 0.505 & 0.330 & 0.105 & 0.152 \\
$\sigma{=}10\%$ & \checkmark & $3,3$ & 16.17 & 0.505 & 0.336 & 0.104 & 0.152 \\
$\sigma{=}20\%$ & \checkmark & $3,3$ & 16.18 & 0.505 & 0.337 & 0.105 & 0.152 \\
\bottomrule
\end{tabular}
\end{table}

\subsection{Distillation into 3DGS}
\label{app:distillation}
Following Difix3D~\cite{wu2025difix3d}, one may distill enhanced close-ups back into the 3DGS. We evaluate three variants: (1) the progressive distillation of Difix3D, (2) that of Difix3D+, and (3) a single distillation step directly from \methodname{}-enhanced close-ups (Table~\ref{tab:distill}). Variants 1 and 2 degrade on all metrics; variant 3 degrades on PSNR, SSIM and LPIPS while slightly improving DINOv2. This mirrors the Difix vs.\ Difix3D trend in Table~\ref{tab:main_results}, although the degradation is milder because scale errors are already mitigated. We attribute this to standard 3DGS fine-tuning being ill-suited to the high-disparity close-up setting; dedicated schemes, such as densifying the close-up region or reweighting close and far camera losses, are a promising direction.

\begin{table}[h]
\centering
\small
\caption{\textbf{Distilling \methodname{} into 3DGS on DL3DV-Closeup.} Variants (1) and (2) follow the distillation of Difix3D and Difix3D+, respectively; variant (3) distills directly from \methodname{}-enhanced close-ups.}
\label{tab:distill}
\begin{tabular}{lccccc}
\toprule
Configuration & PSNR $\uparrow$ & SSIM $\uparrow$ & LPIPS $\downarrow$ & DreamSim $\downarrow$ & DINOv2 $\downarrow$ \\
\midrule
\methodname{} (no distillation) & \textbf{16.17} & \textbf{0.505} & \textbf{0.330 }& \textbf{0.105 }& 0.152 \\
Distillation (variant 1) & 15.56 & 0.469 & 0.366 & 0.116 & 0.177 \\
Distillation (variant 2) & 15.32 & 0.454 & 0.356 & 0.117 & 0.168 \\
Distillation (variant 3) & 15.81 & 0.484 & 0.339 & \textbf{0.105} & \textbf{0.146} \\
\bottomrule
\end{tabular}
\end{table}

\newpage
\section{Benchmark Construction Details}
\label{sec:benchmark_details}
We provide additional details for the close-up novel view synthesis benchmarks used in our experiments. We first describe the construction of the DL3DV-Closeup benchmark in Appendix \ref{sec:dl3dv_disparity}, followed by MobileClose-10 (Appendix \ref{sec:self_collected_benchmark}), which was captured specifically for this task.

\paragraph{Zoom statistics.} Estimating the effective zoom factor for closeups directly from COLMAP or 3DGS-rendered depth is challenging due to sparsity and noise. We therefore use MASt3R~\cite{mast3r} depth estimation for the scene (including the closeup GT images), which provides cleaner depth. We then estimate the effective zoom factor as $\text{Zoom} = \lVert \mathbf{c}_W - \mathbf{P} \rVert / \lVert \mathbf{c}_C - \mathbf{P} \rVert$, where $\mathbf{c}_C$ and $\mathbf{c}_W$ are the closeup and wide camera centers and $\mathbf{P}$ is the median of the 3D points unprojected from the closeup that fall inside the wide view. Across DL3DV-Closeup we measure $\text{Zoom} = \times 2.59 \pm 0.73$ (range $\times1.46$-$\times5.81$), and across MobileClose-10 $\text{Zoom} = \times3.24 \pm 1.03$ (range $\times1.47$-$\times5.16$), confirming that both benchmarks contain genuine closeups with substantial scale gap from the training views.

\paragraph{Exposure differences.} In some scenes from both benchmarks, the close-up and wide views were captured under slightly different exposure settings. Reference-conditioned enhancers, including MACRO, inherit the wide-shot's exposure when transferring appearance from the reference. This caps the achievable PSNR and SSIM on these scenes for all reference-conditioned methods we evaluate, and is one reason the absolute reconstruction scores in Table \ref{tab:main_results} are modest. Since all enhancement baselines are affected equally, the comparison between methods remains fair.

\subsection{DL3DV-Closeup Benchmark}
\label{sec:dl3dv_disparity}
We construct DL3DV-Closeup from the DL3DV-10K benchmark~\cite{ling2024dl3dv}, which contains 140 scenes. We filter scenes according to depth disparity and valid close-up/far-view relationships, resulting in 44 scenes and a total of 314 close-up evaluation images. We show examples of far and close-up views in Appendix \ref{sec:dl3dv_disparity_examples}, and describe the construction of the benchmarks in Appendices \ref{sec:per_camera_depth_estimation}- \ref{sec:train_test_split}.

\subsubsection{Dataset Examples}
\label{sec:dl3dv_disparity_examples}

\begin{figure*}[h]
    \centering
    \setlength{\tabcolsep}{0pt}
    \renewcommand{\arraystretch}{0}
    \begin{tabular}{@{}cccccc@{}}
        \includegraphics[width=0.165\linewidth]{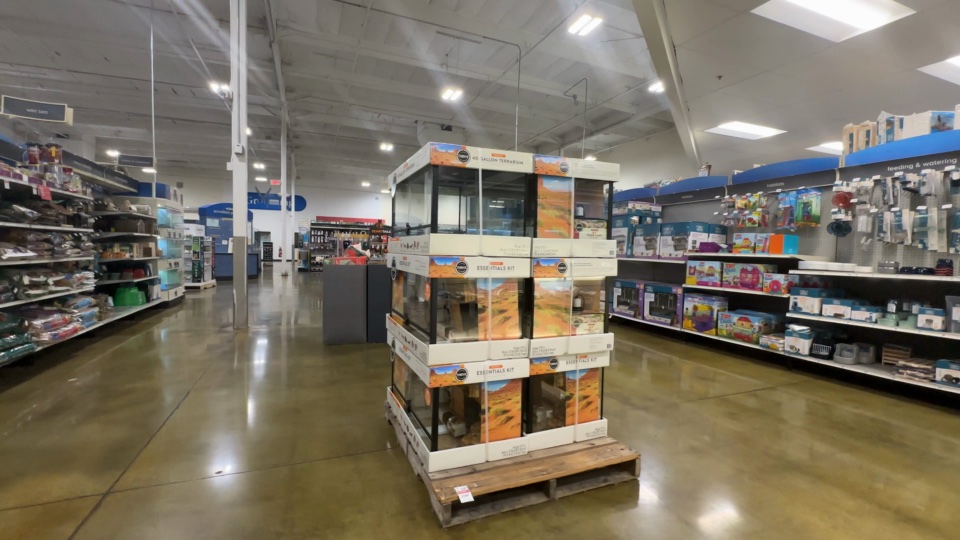} &
        \includegraphics[width=0.165\linewidth]{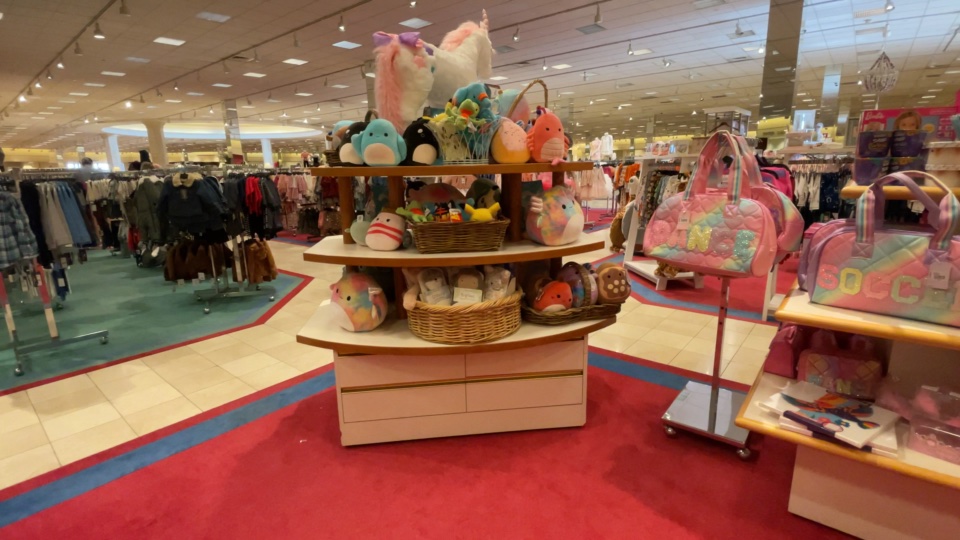} &
        \includegraphics[width=0.165\linewidth]{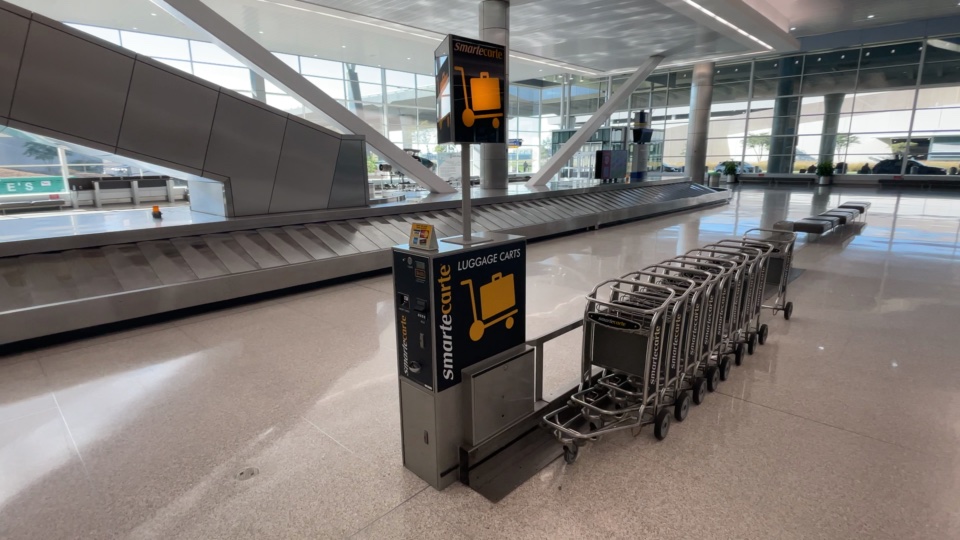} &
        \includegraphics[width=0.165\linewidth]{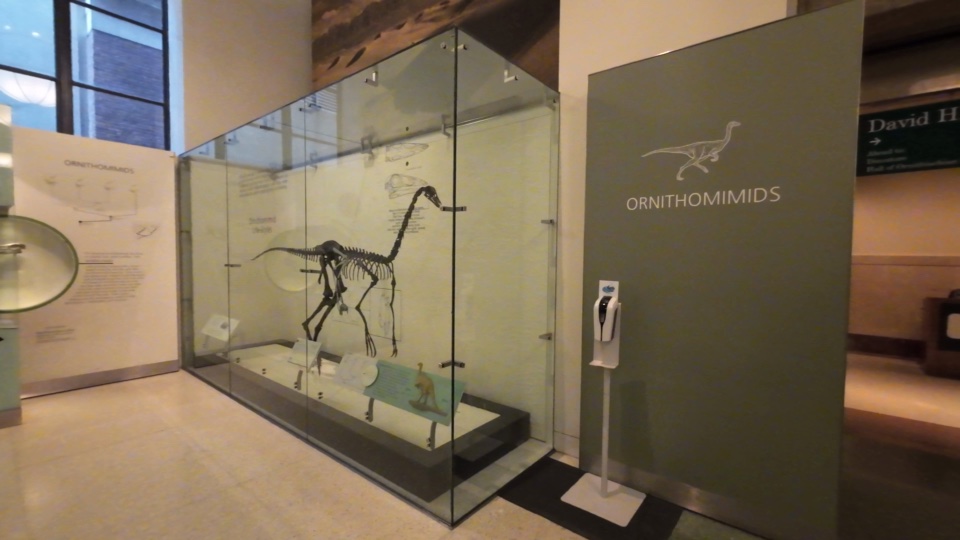} &
        \includegraphics[width=0.165\linewidth]{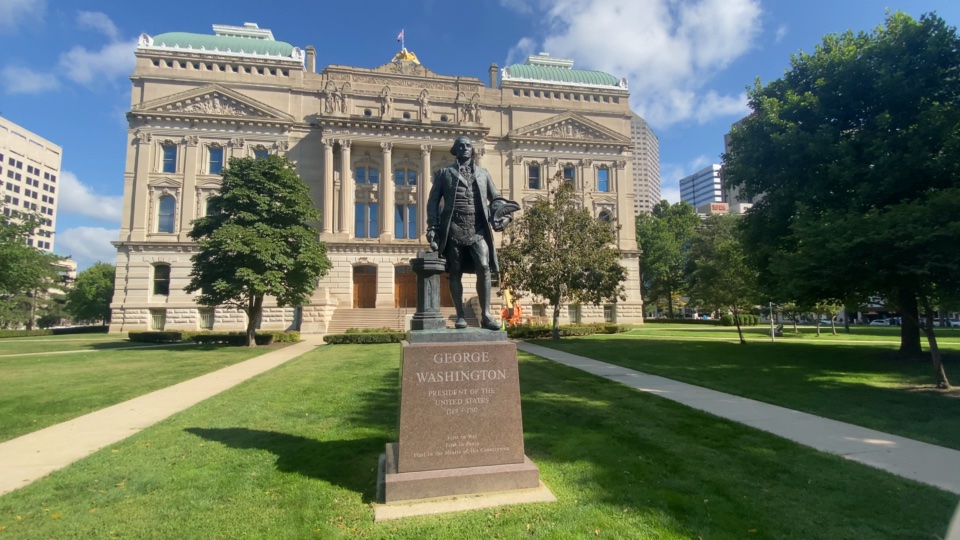} &
        \includegraphics[width=0.165\linewidth]{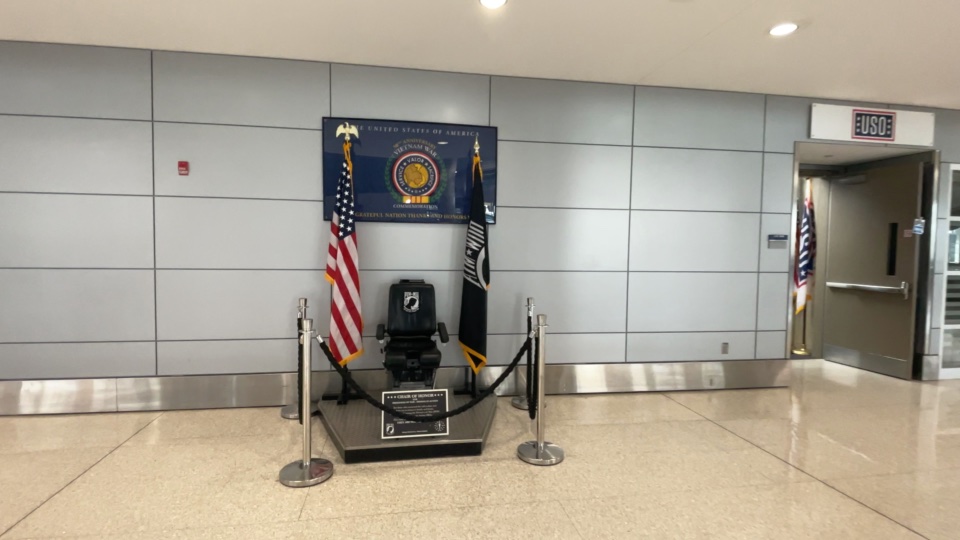} \\
        \includegraphics[width=0.165\linewidth]{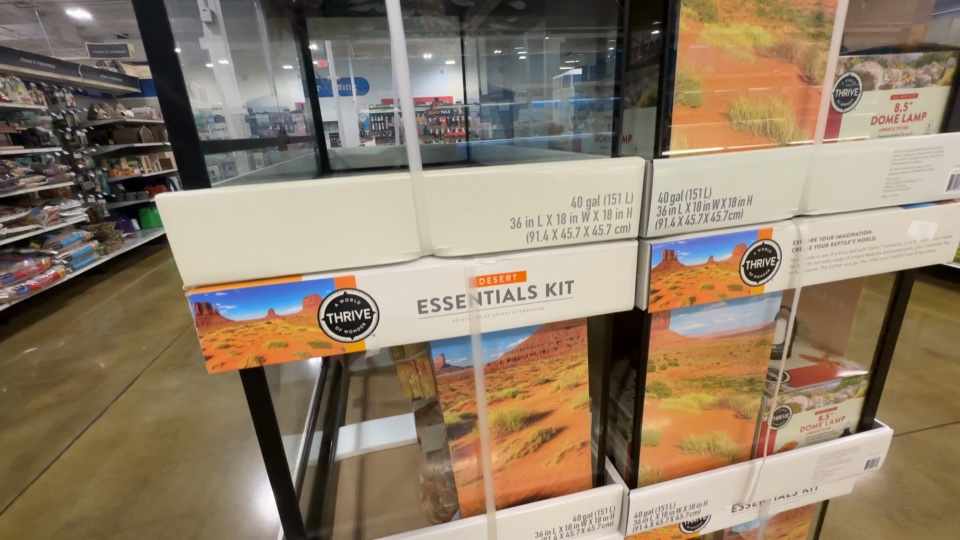} &
        \includegraphics[width=0.165\linewidth]{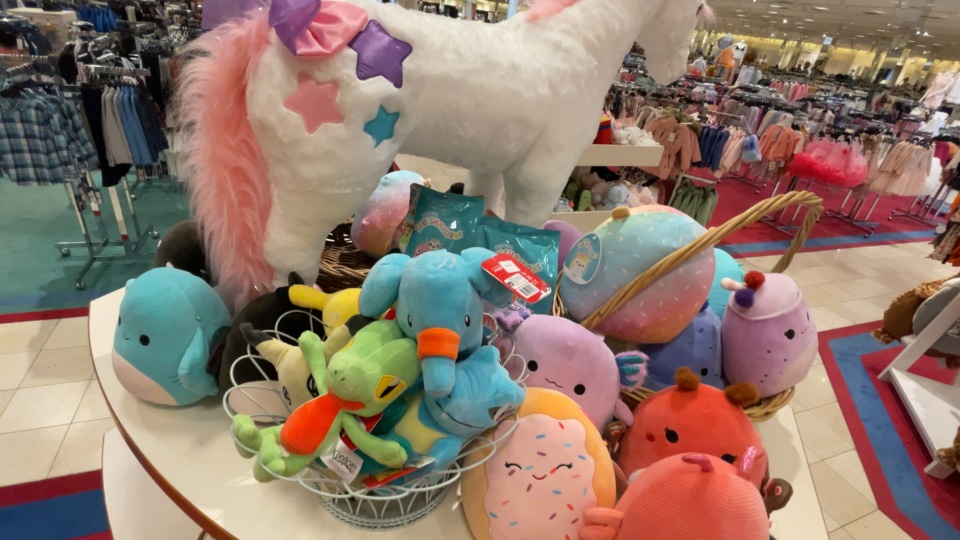} &
        \includegraphics[width=0.165\linewidth]{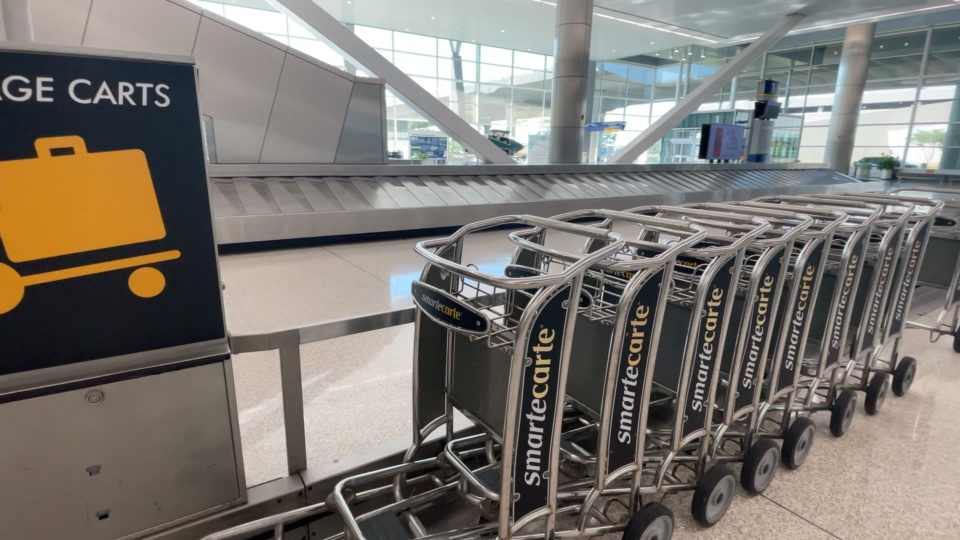} &
        \includegraphics[width=0.165\linewidth]{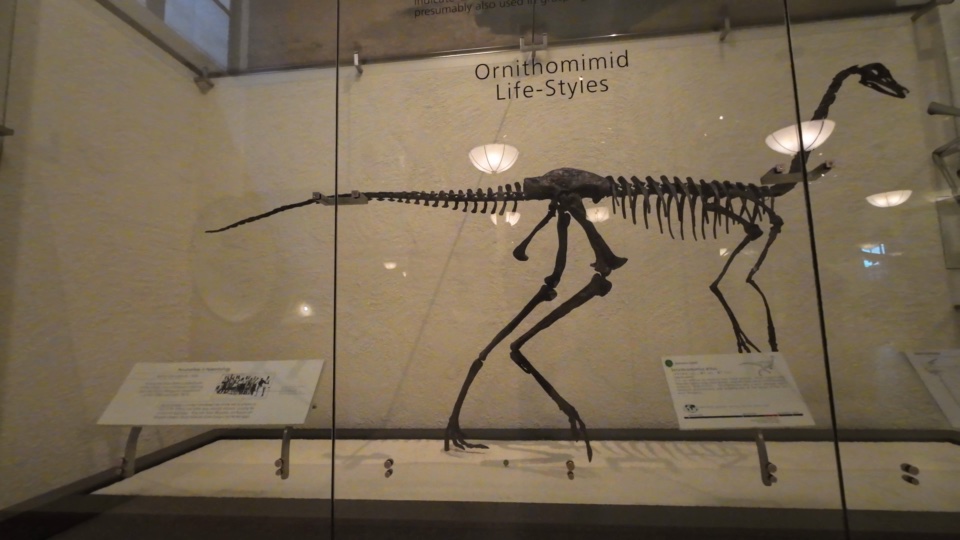} &
        \includegraphics[width=0.165\linewidth]{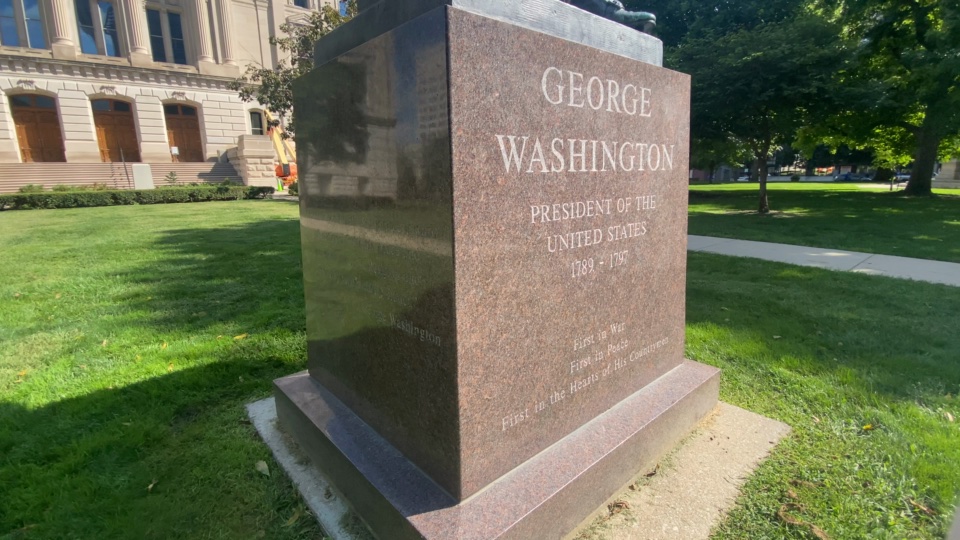} &
        \includegraphics[width=0.165\linewidth]{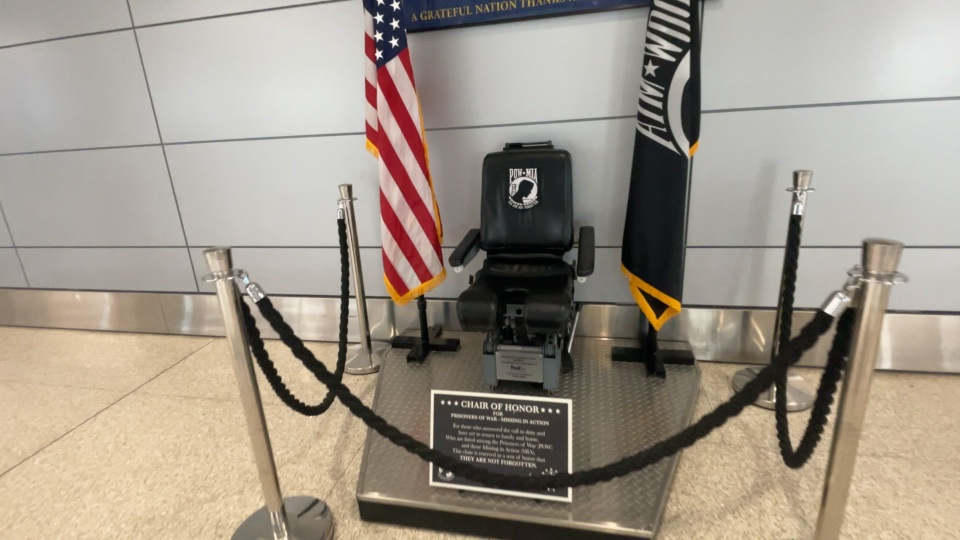} \\[4pt]
        \includegraphics[width=0.165\linewidth]{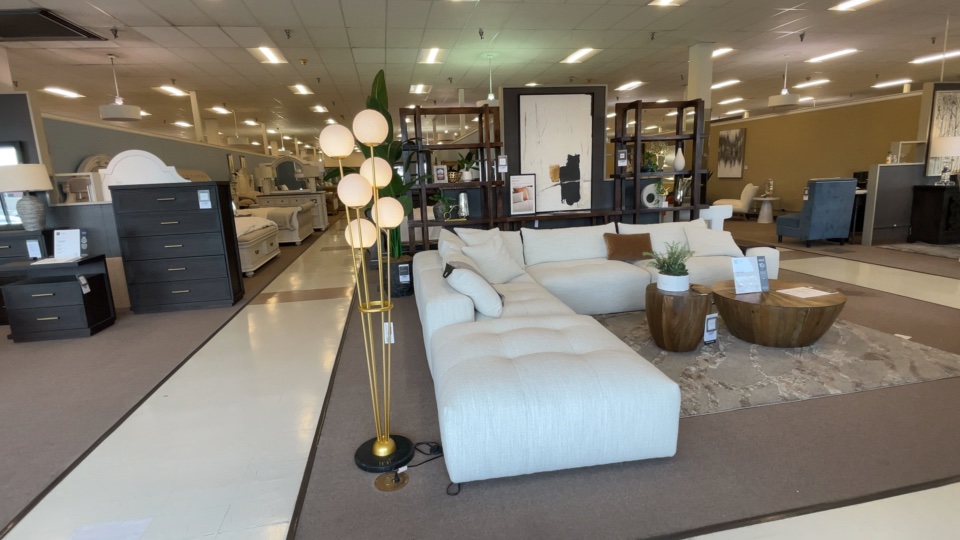} &
        \includegraphics[width=0.165\linewidth]{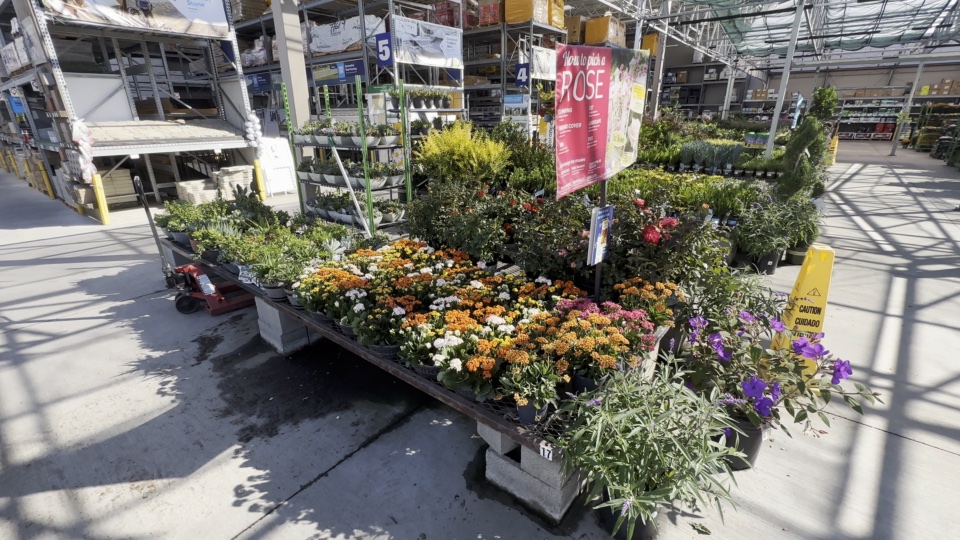} &
        \includegraphics[width=0.165\linewidth]{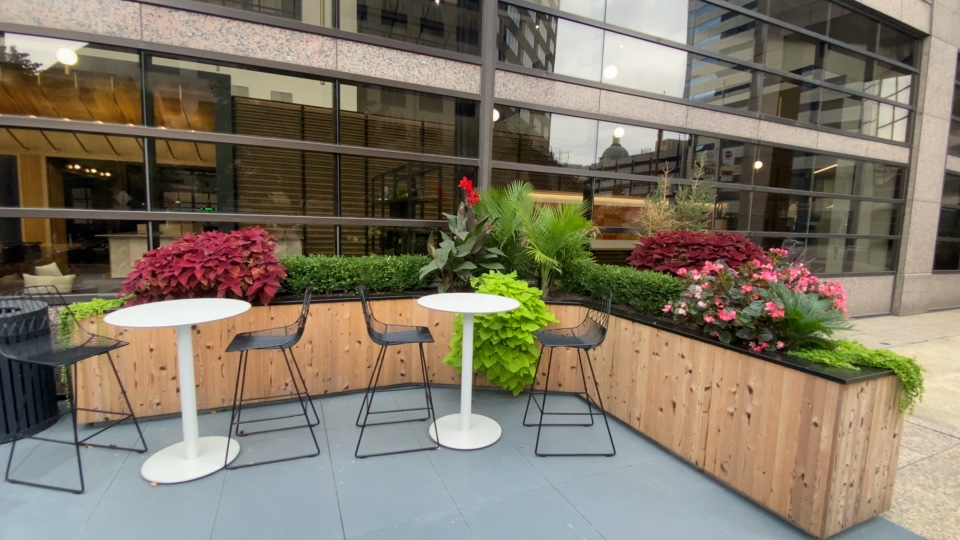} &
        \includegraphics[width=0.165\linewidth]{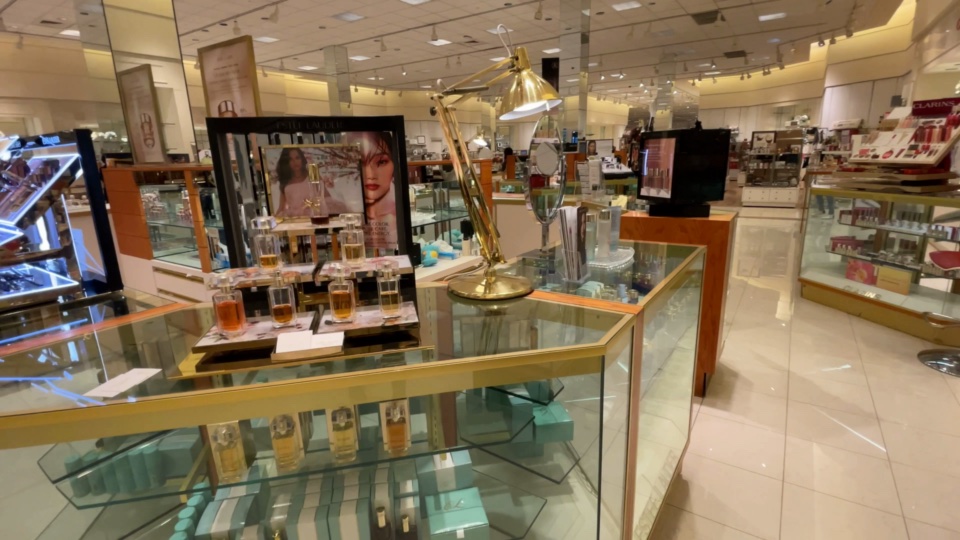} &
        \includegraphics[width=0.165\linewidth]{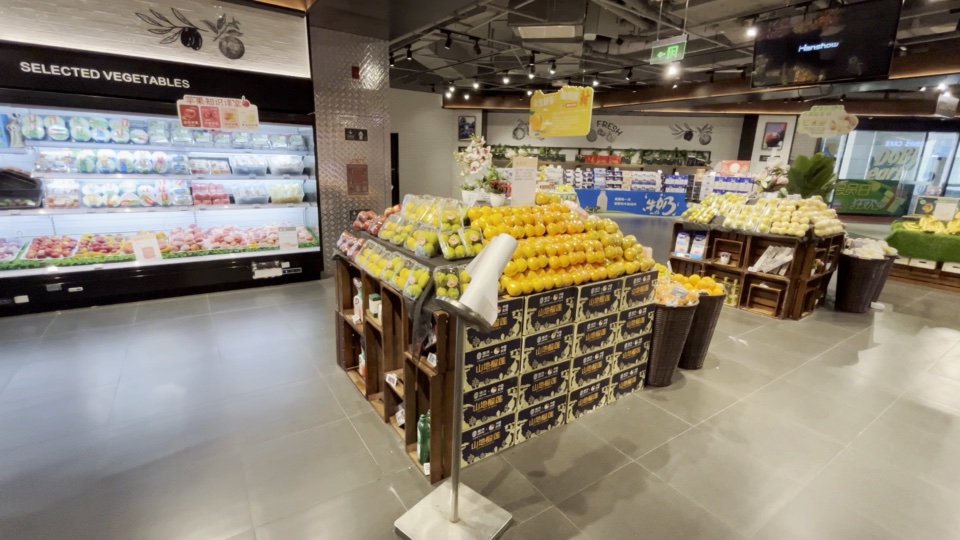} &
        \includegraphics[width=0.165\linewidth]{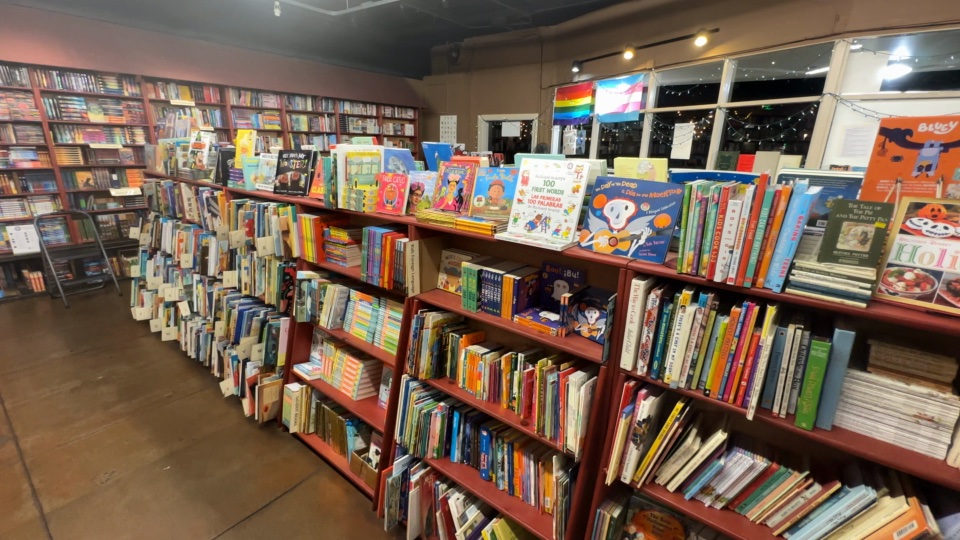} \\
        \includegraphics[width=0.165\linewidth]{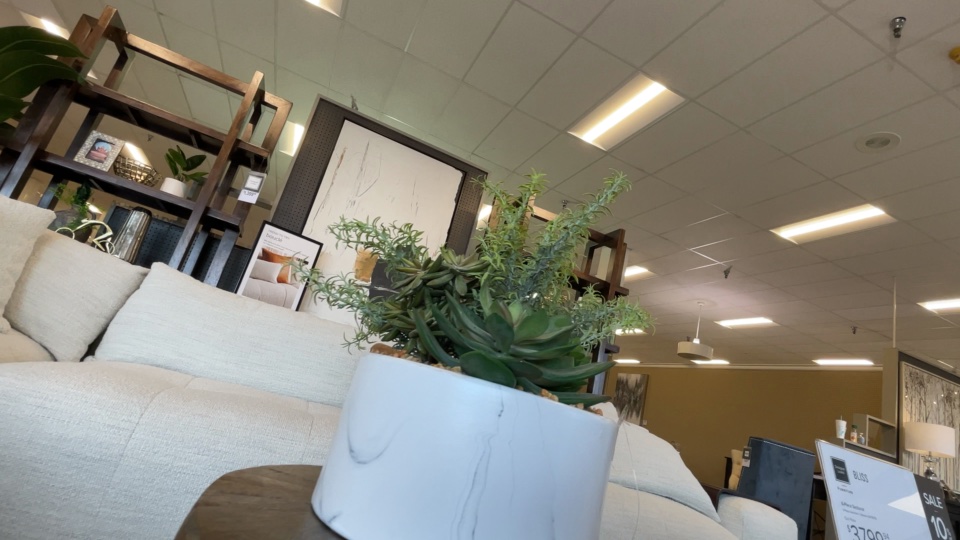} &
        \includegraphics[width=0.165\linewidth]{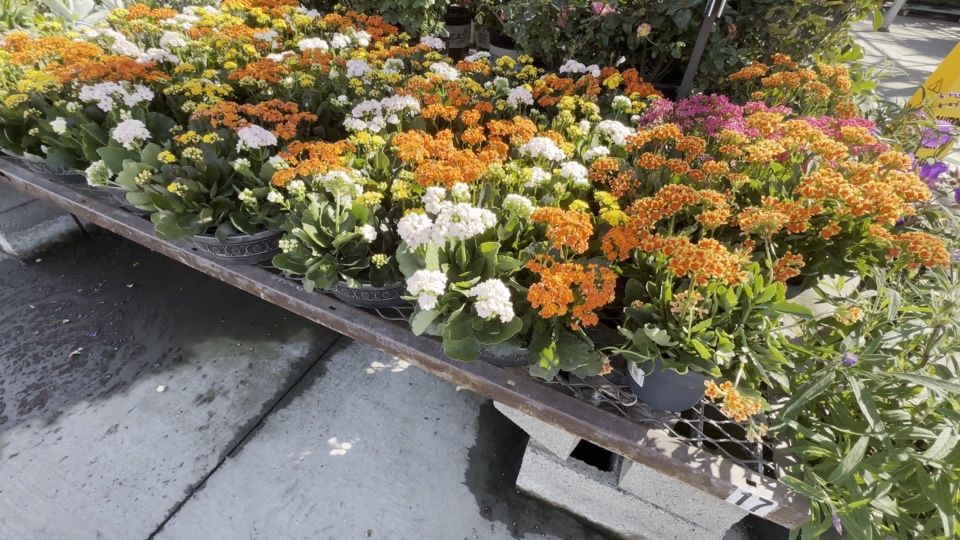} &
        \includegraphics[width=0.165\linewidth]{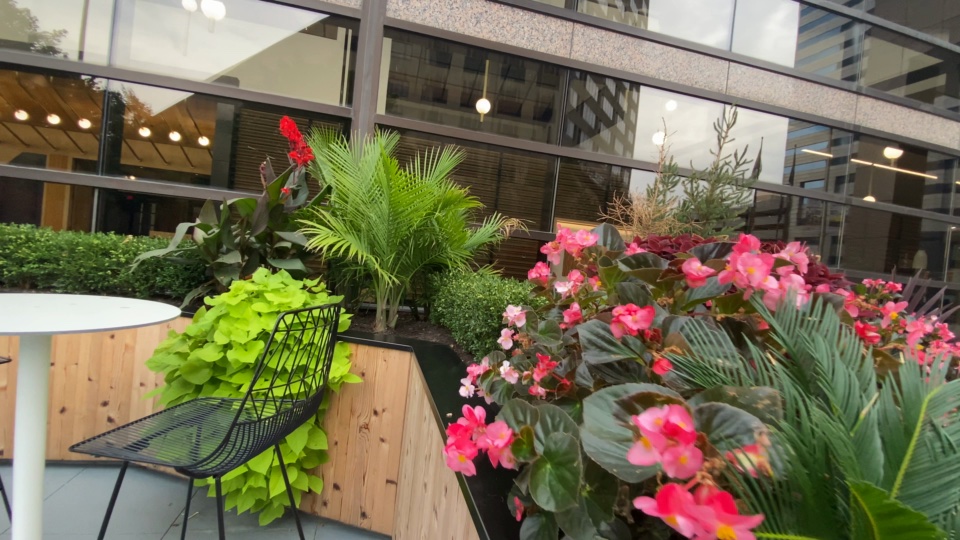} &
        \includegraphics[width=0.165\linewidth]{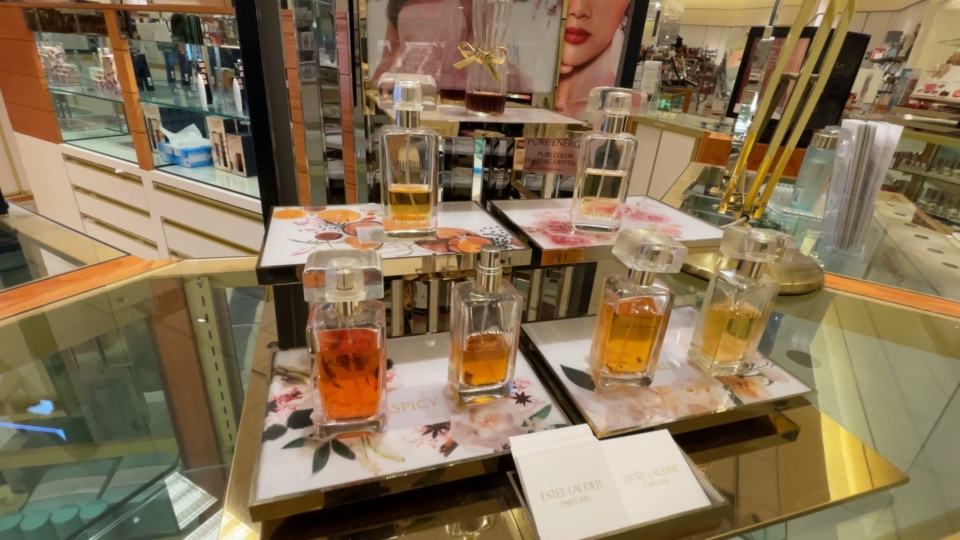} &
        \includegraphics[width=0.165\linewidth]{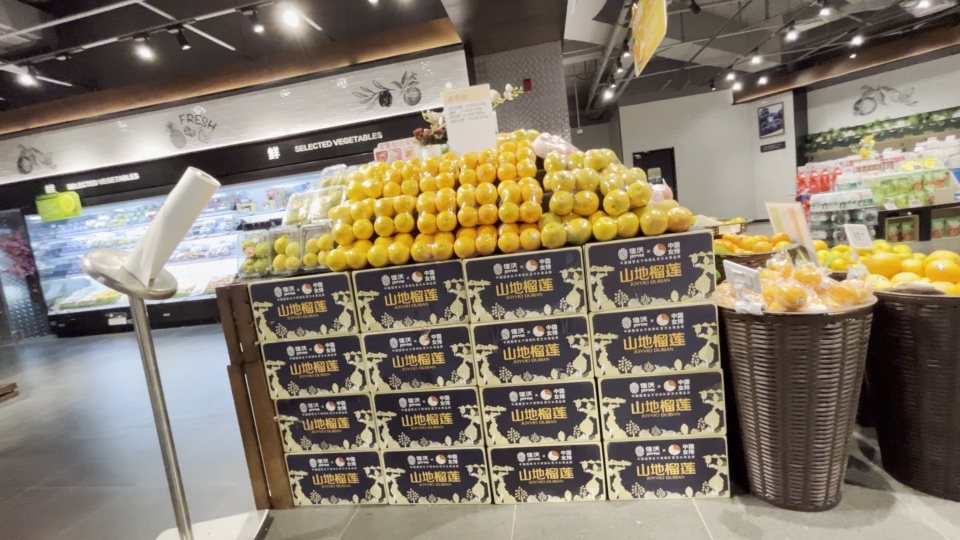} &
        \includegraphics[width=0.165\linewidth]{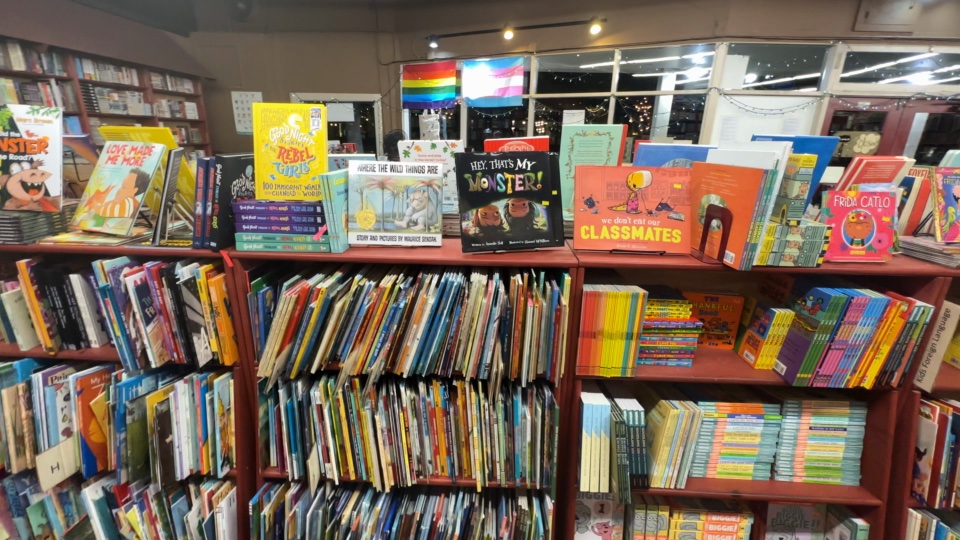}
    \end{tabular}
    \caption{\textbf{Example scenes from DL3DV-Closeup.} Odd rows show far training views, even rows show corresponding close-up evaluation views. Each column is one scene. Best viewed zoomed-in on a digital screen.}
    \label{fig:dl3dv_disparity_examples}
\end{figure*}

Figure \ref{fig:dl3dv_disparity_examples} shows example scenes from our DL3DV-Closeup benchmark. Each example includes far training views and corresponding close-up evaluation views, illustrating the large scale changes and partial-view setting targeted by the benchmark. 

\subsubsection{Per-Camera Depth Estimation}
\label{sec:per_camera_depth_estimation}

For each scene, we use the COLMAP sparse reconstruction $\mathcal{X} = \{x_m\}_{m=1}^{M}$ and camera poses $\{(R_i, t_i)\}_{i=1}^{N}$. Each camera $\pi_i$ is represented by its pose $(R_i, t_i)$ and viewing direction $\mathbf{v}_i$, where $\mathbf{v}_i$ denotes the unit optical axis.

We define the set of visible points for camera $i$ as:
\begin{equation}
\mathcal{V}_i = \left\{ x_m \in \mathcal{X} \;\middle|\;
\frac{\langle x_m - t_i,\ \mathbf{v}_i \rangle}{\|x_m - t_i\|} > \cos\left(\tfrac{\theta_{\text{fov}}}{2}\right),\;
\langle x_m - t_i,\ \mathbf{v}_i \rangle > 0
\right\},
\end{equation}
where $\theta_{\text{fov}}$ is the diagonal field of view derived from the camera intrinsics. 
This corresponds to 3D points within the camera's field of view and in front of the camera, providing a coarse approximation of visible scene content.

We define the \emph{median scene depth} for camera $i$ as:
\begin{equation}
d_i = \mathrm{median}_{x_m \in \mathcal{V}_i} \ \langle x_m - t_i,\ \mathbf{v}_i \rangle.
\end{equation}
This corresponds to the median depth of visible content along the camera viewing direction.

We further compute the 10th and 90th percentile depths, $d_i^{10}$ and $d_i^{90}$, and define the within-view depth span as $s_i = d_i^{90} / d_i^{10}$, capturing the range of depths observed within a single view.

\subsubsection{Candidate Pair Construction}
\label{sec:candidate_pair_construction}

For each scene, cameras are partitioned into a \emph{close} set (bottom 40\% by $d_i$) and a \emph{far} set (top 40\% by $d_i$). 

We define candidate close-up/far pairs $(i,j)$, where $i$ is a close camera and $j$ is a far camera. A pair must satisfy:
\begin{itemize}
    \item Depth ratio $\frac{d_j}{d_i} \geq 2$,
    \item Intersection-over-Union (IoU) $\geq 40\%$, computed between the sets of visible 3D points $\mathcal{V}_i$ and $\mathcal{V}_j$,
    \item Viewing angle difference between $10^\circ$ and $60^\circ$.
\end{itemize}

Let $N_{\text{cand}}$ denote the number of such valid candidate pairs in the scene.

\subsubsection{Scene Selection}
\label{sec:scene_selection}

We rank scenes using a disparity score that captures both cross-view and within-view depth variation:
\begin{equation}
\resizebox{0.95\linewidth}{!}{$
S = \frac{\sigma(\{d_i\})}{\mu(\{d_i\})}
\cdot \left(1 + \log(1 + \bar{s})\right)
\cdot \left(1 + \log\left(1 + \frac{\max_i d_i^{10}}{\min_i d_i^{10}}\right)\right)
\cdot \left(1 + \log\left(1 + \frac{1}{\min_i d_i}\right)\right)
\cdot \log(1 + N_{\text{cand}})
$}
\end{equation}
where $\bar{s} = \frac{1}{N} \sum_i s_i$, and $\sigma(\cdot)$ and $\mu(\cdot)$ denote standard deviation and mean, respectively.

The score is composed of the following factors:
\begin{itemize}
    \item \textbf{Cross-view depth variation:} 
    $\frac{\sigma(\{d_i\})}{\mu(\{d_i\})}$ measures the coefficient of variation of per-camera depths, capturing how much scene depth varies across viewpoints.

    \item \textbf{Within-view depth span:} 
    $\bar{s}$ summarizes the depth range within individual views, where $s_i = d_i^{90} / d_i^{10}$.

    \item \textbf{Near-depth variation:} 
    $\frac{\max_i d_i^{10}}{\min_i d_i^{10}}$ captures variation in the closest visible content across views.

    \item \textbf{Closeness:} 
    $\min_i d_i$ encourages the presence of cameras observing very close scene content.

    \item \textbf{Pair count:} 
    $N_{\text{cand}}$ denotes the number of valid candidate close-up/far pairs, ensuring that meaningful pairs exist in the scene.
\end{itemize}

Logarithmic scaling is applied to stabilize the contribution of each term and prevent extreme values from dominating the score.

Scenes with $N_{\text{cand}} = 0$ are assigned a score of zero and discarded.
\subsubsection{Train/Test Split and Evaluation Pairs}
\label{sec:train_test_split}
For each selected scene, frames are sorted by their median depth $d_i$ and divided as follows:
\begin{itemize}
    \item Top 30\% (farthest views): training pool,
    \item Bottom 30\% (closest views): test candidates.
\end{itemize}
The middle 40\% is discarded to enforce a clear separation between training and test views.

From the training pool, we select $N=16$ views using farthest-point sampling over camera positions to maximize spatial coverage of the scene geometry.

\subsection{MobileClose-10 Benchmark}
\label{sec:self_collected_benchmark}

In addition to DL3DV-Closeup, we introduce MobileClose-10, a close-up benchmark captured specifically for this task. The dataset was collected using an iPhone 17 Pro Max across 10 diverse real-world scenes.

\begin{figure*}[h]
    \centering
    \setlength{\tabcolsep}{0pt}
    \renewcommand{\arraystretch}{0}
    \resizebox{\textwidth}{!}{%
    \begin{tabular}{@{}ccccccccc@{}}
        \includegraphics[width=0.11\linewidth]{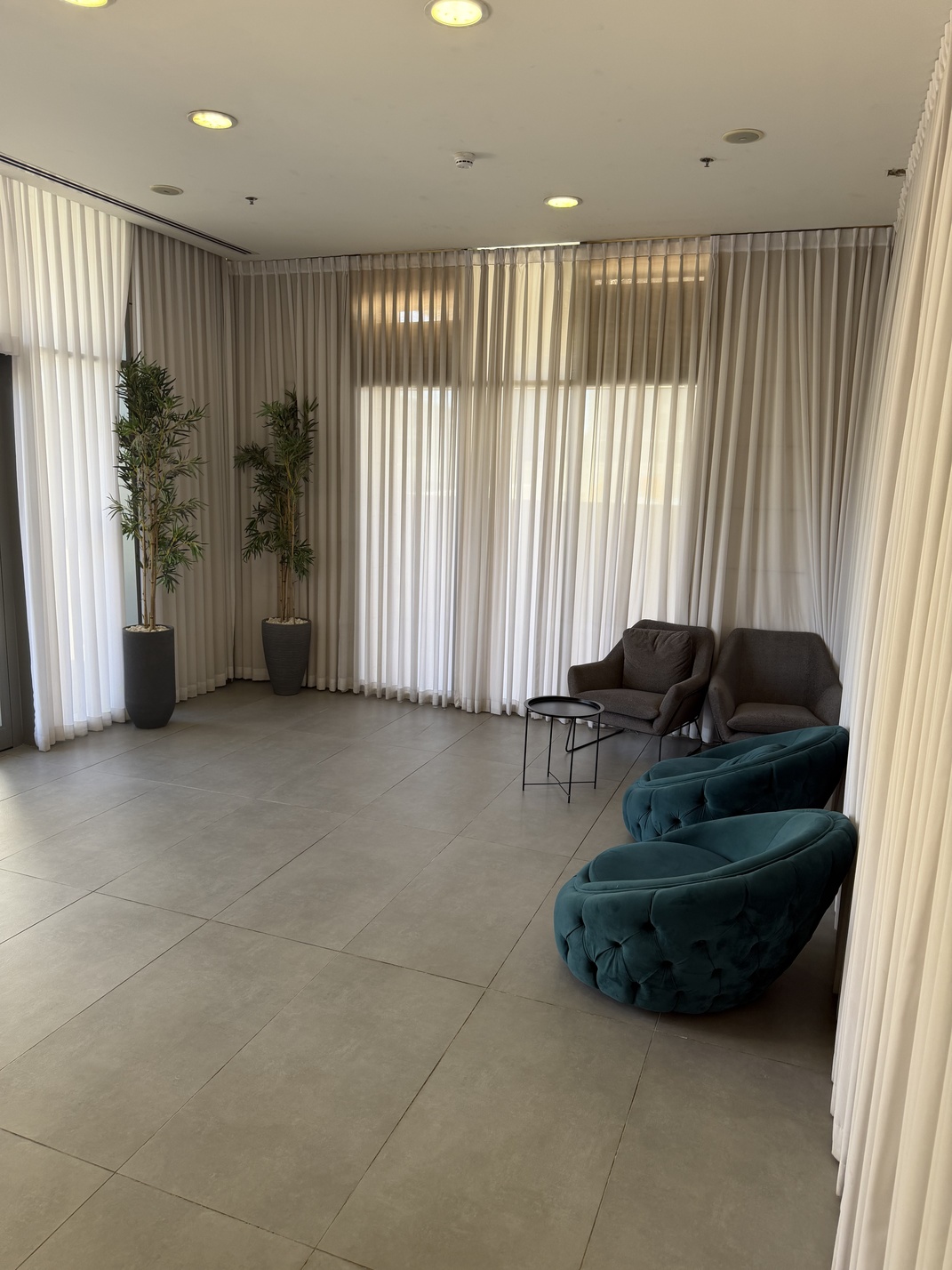} &
        \includegraphics[width=0.11\linewidth]{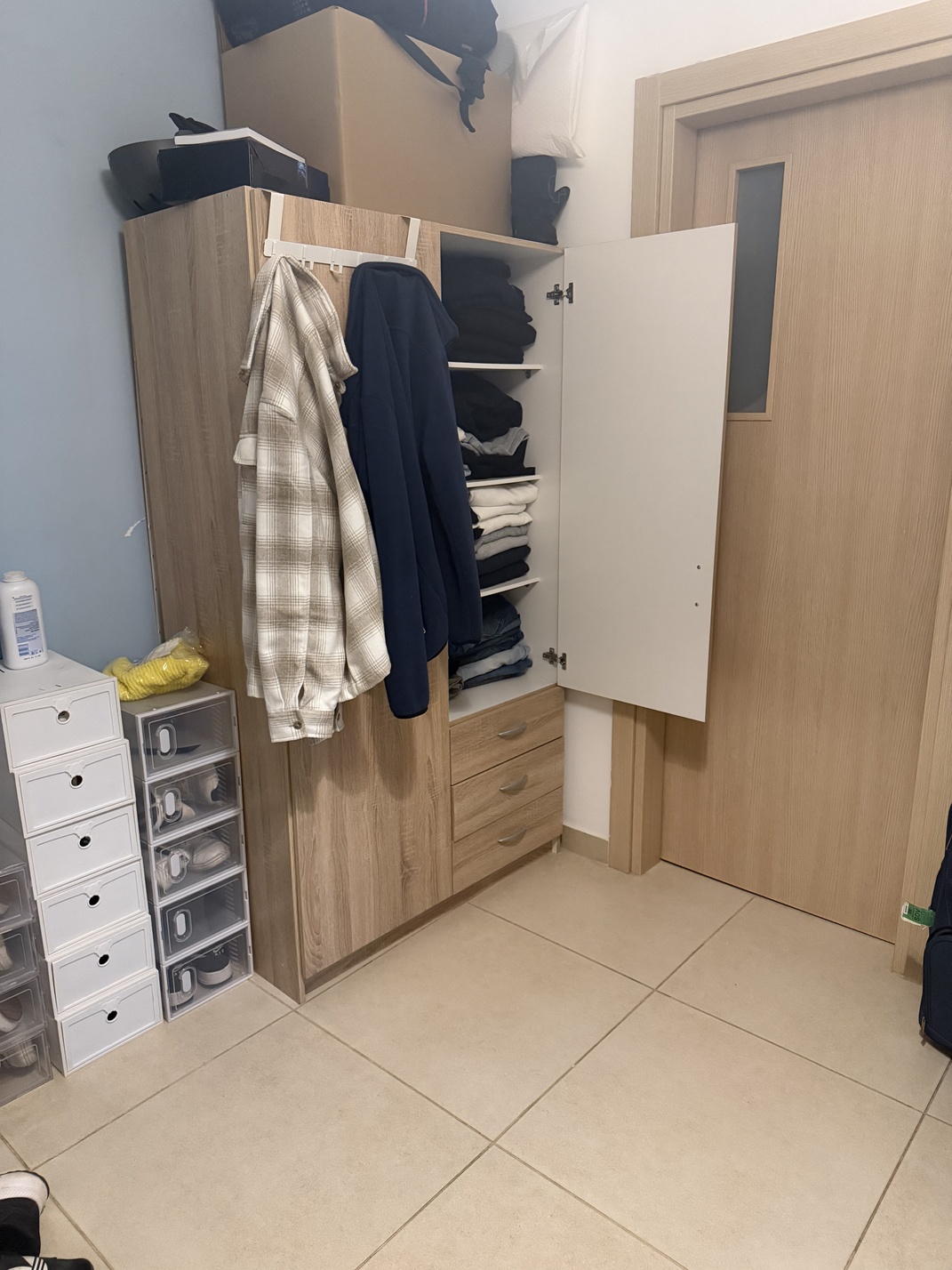} &
        \includegraphics[width=0.11\linewidth]{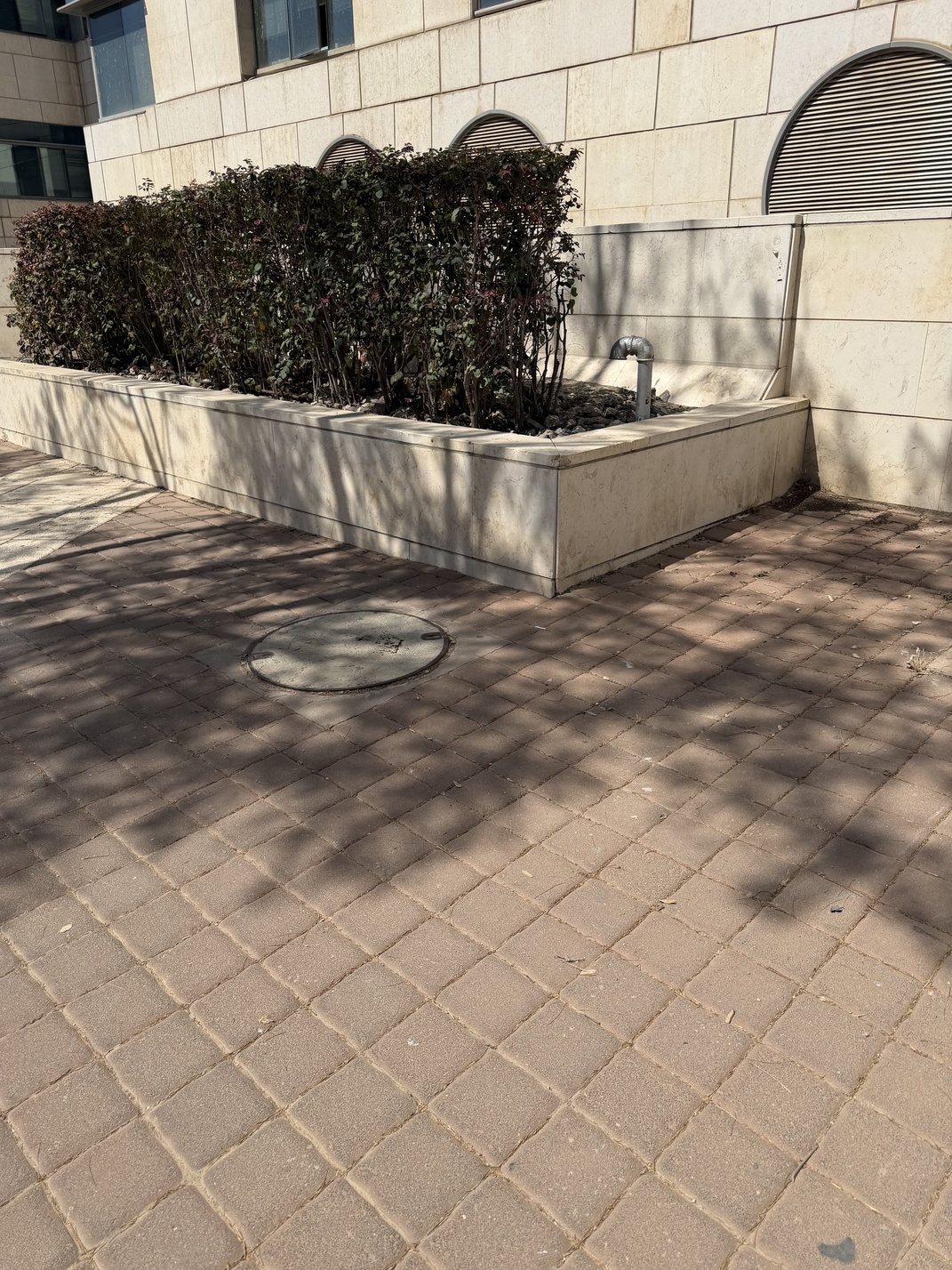} &
        \includegraphics[width=0.11\linewidth]{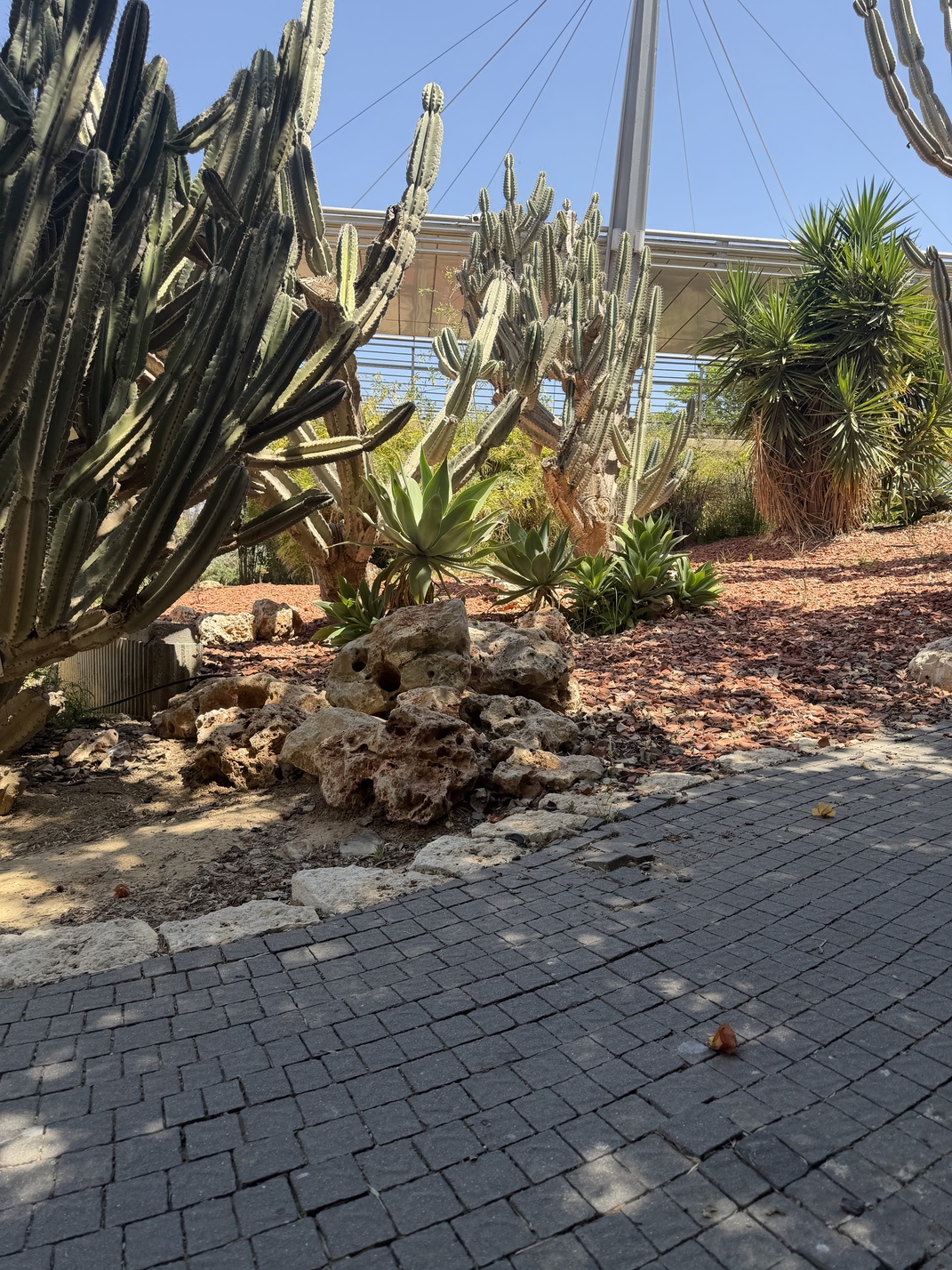} &
        \includegraphics[width=0.11\linewidth]{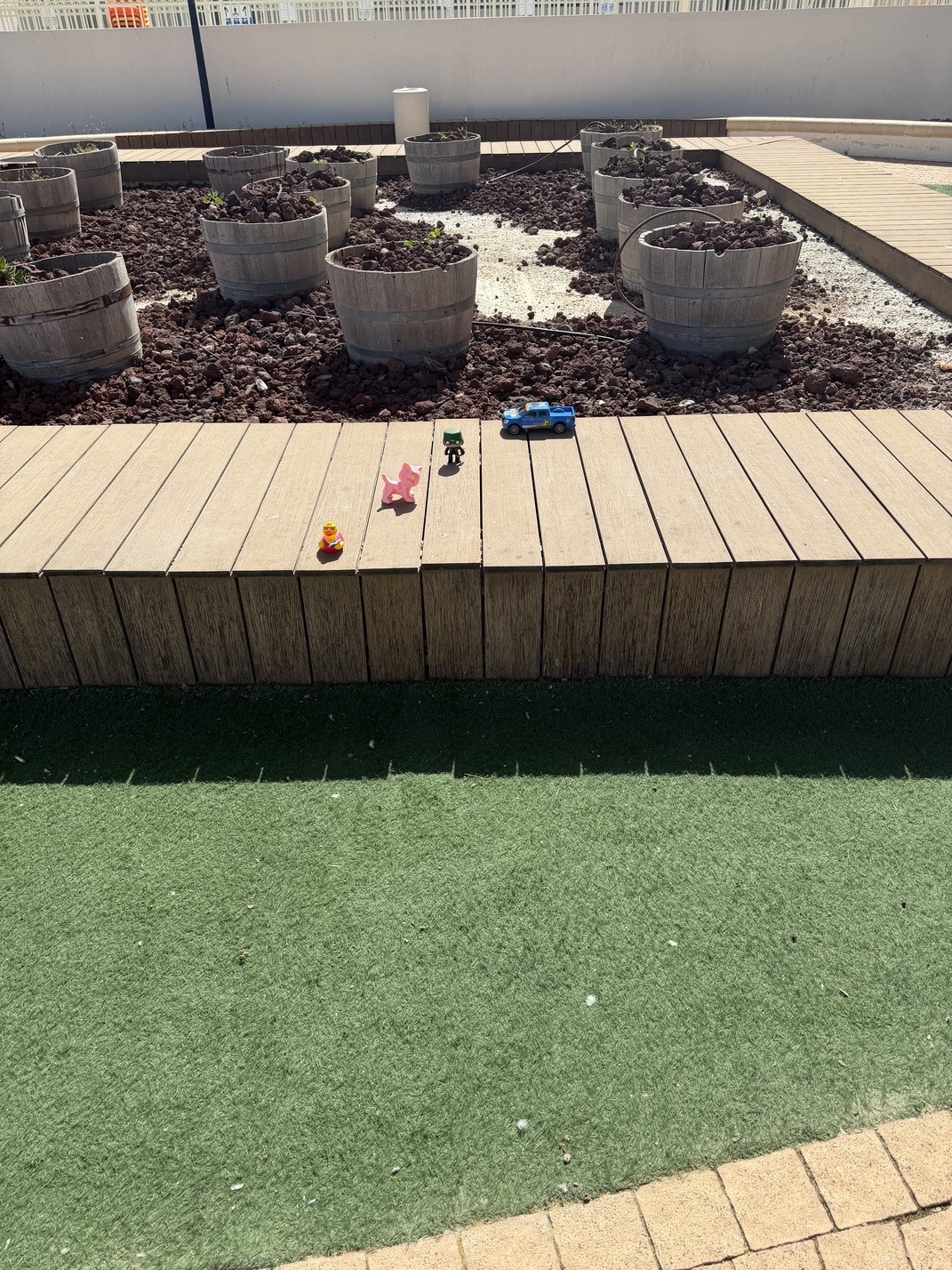} &
        \includegraphics[width=0.11\linewidth]{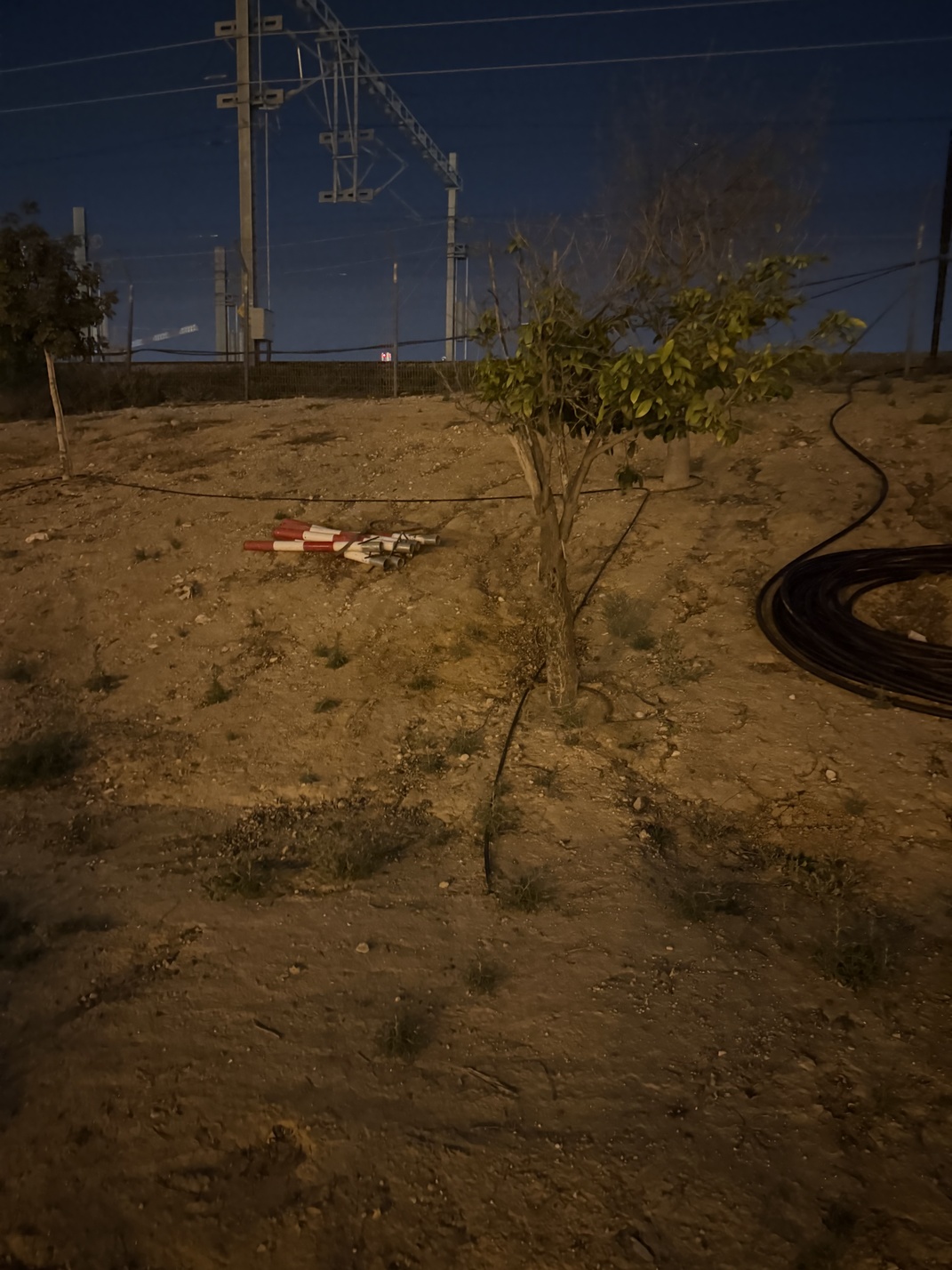} &
        \includegraphics[width=0.11\linewidth]{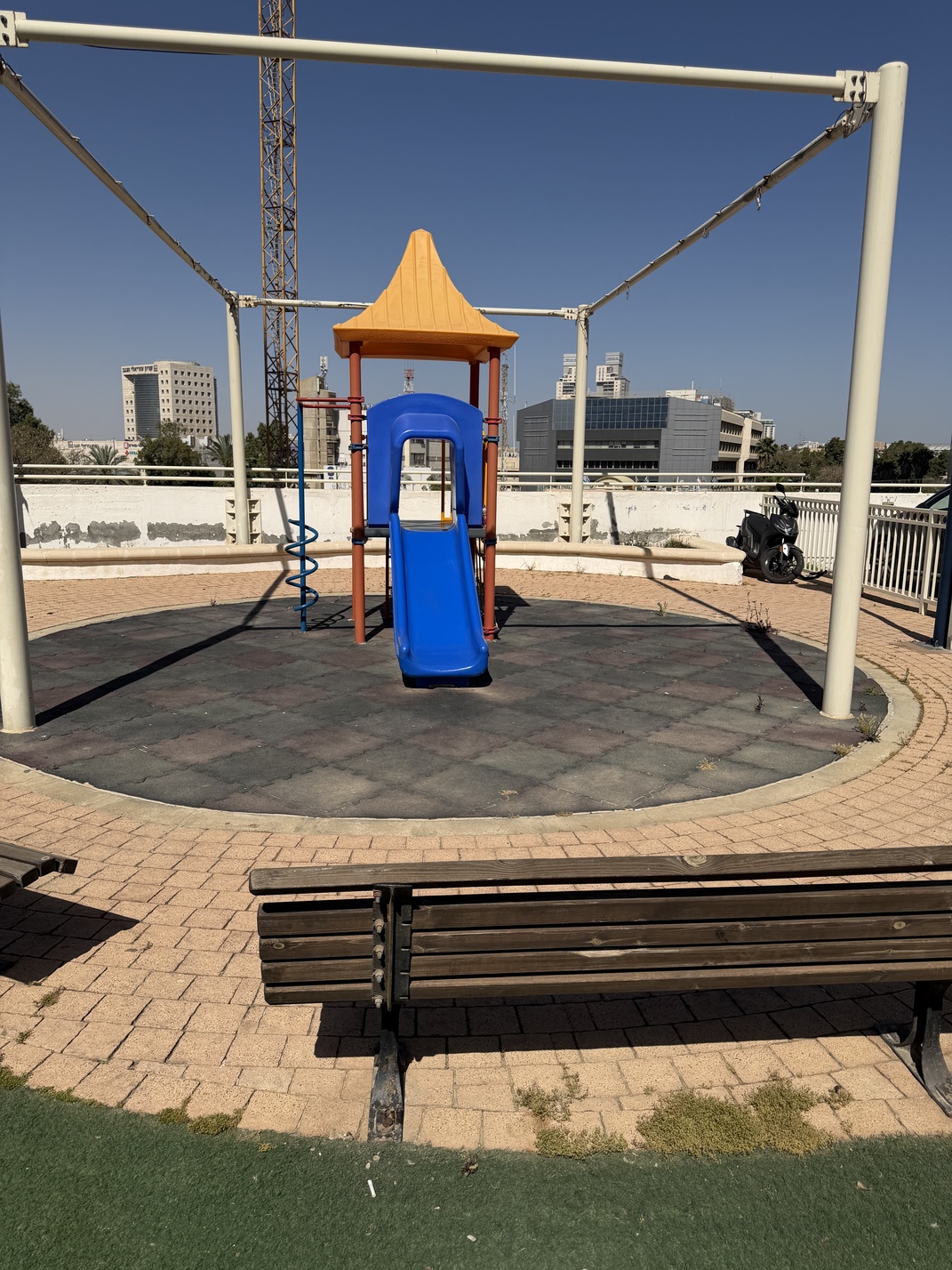} &
        \includegraphics[width=0.11\linewidth]{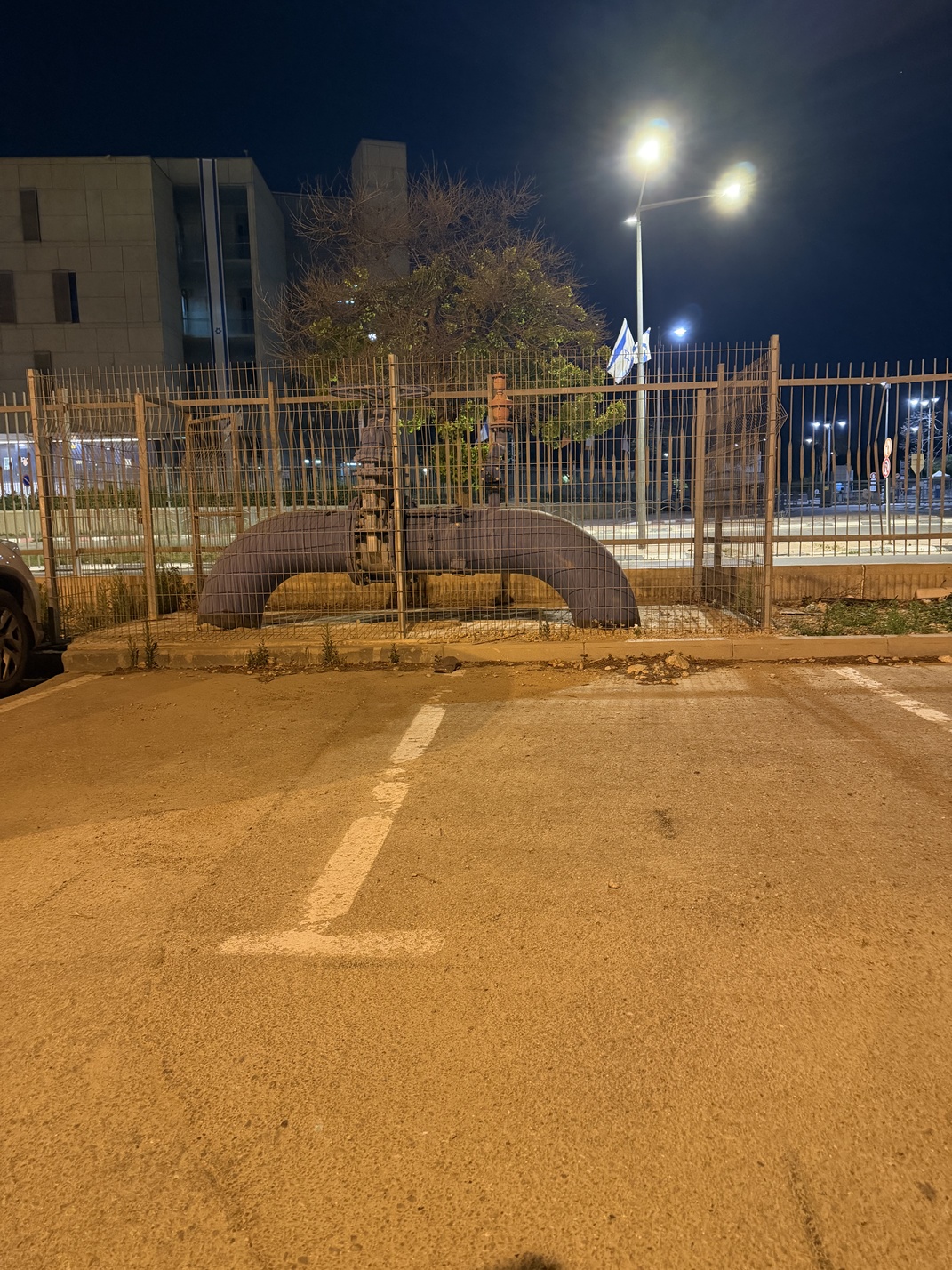} &
        \includegraphics[width=0.11\linewidth]{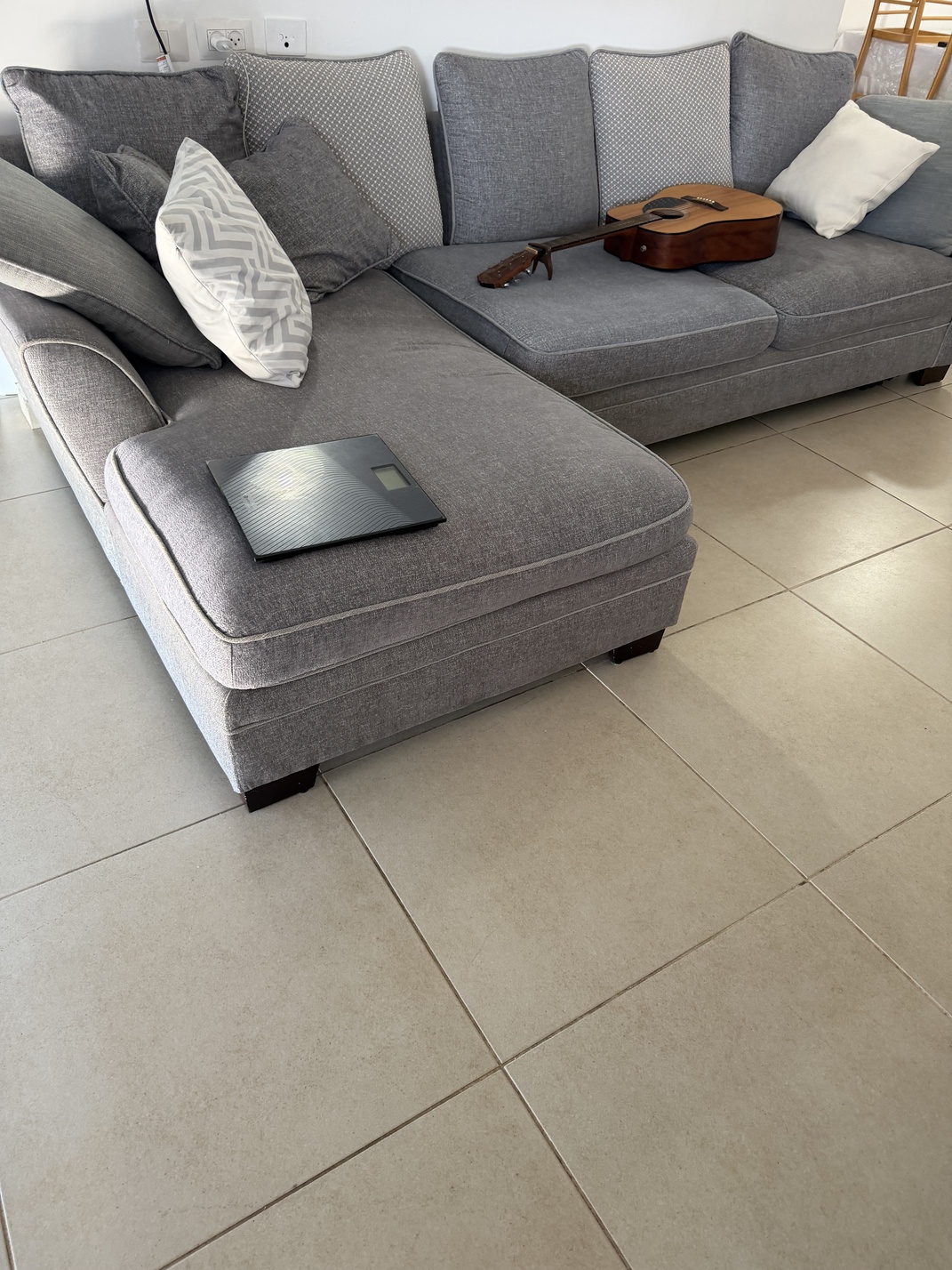} \\

        \includegraphics[width=0.11\linewidth]{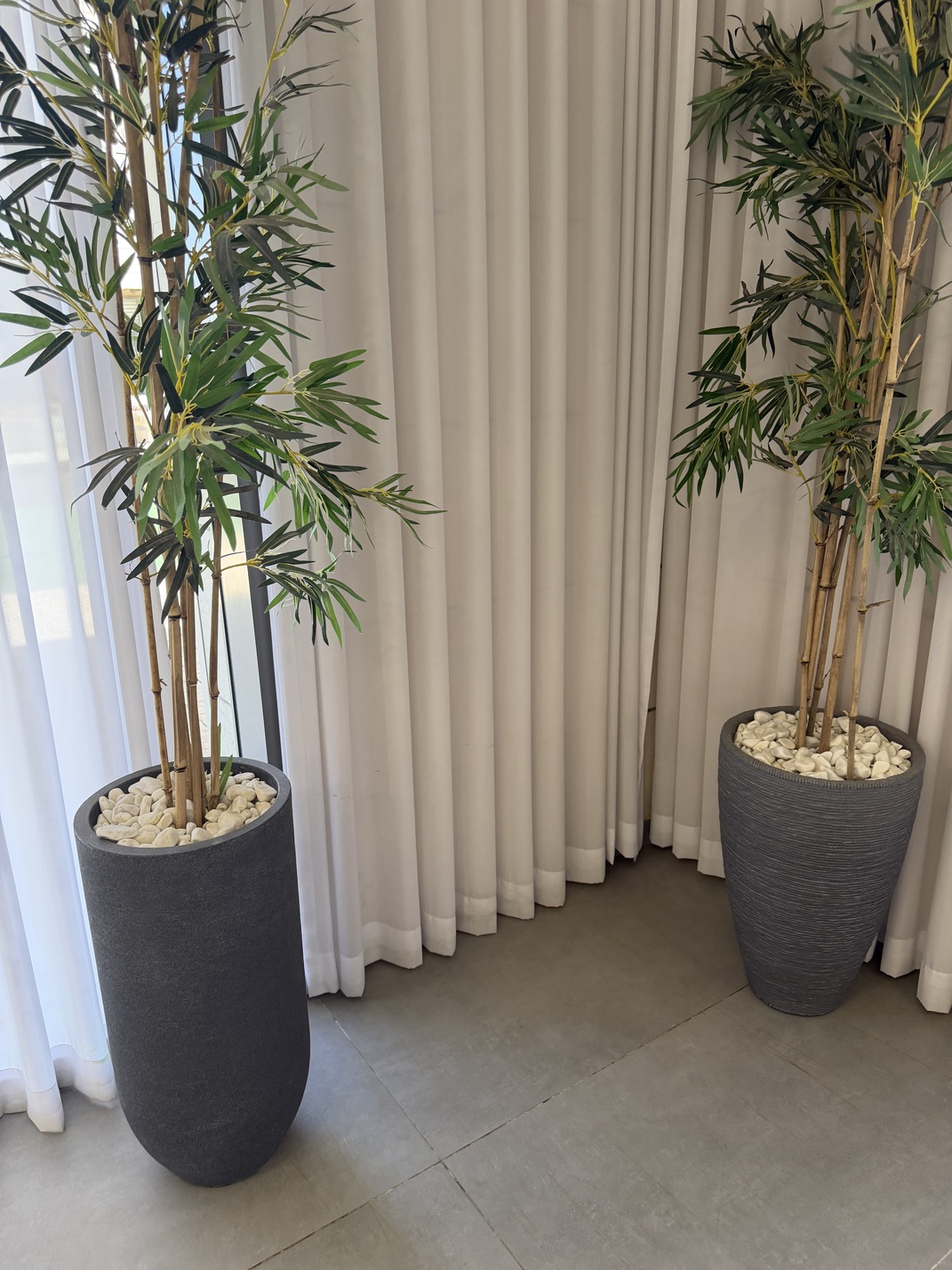} &
        \includegraphics[width=0.11\linewidth]{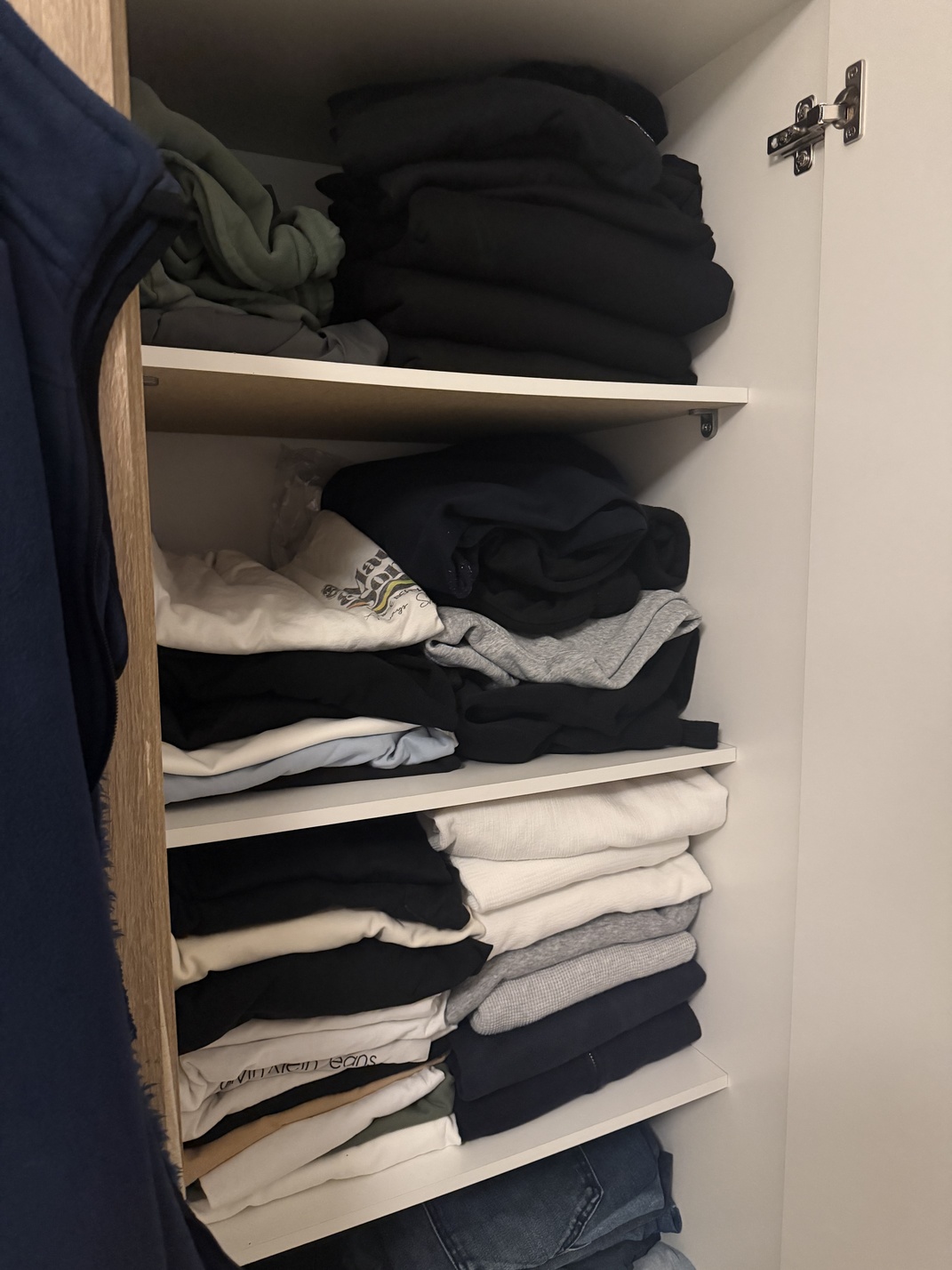} &
        \includegraphics[width=0.11\linewidth]{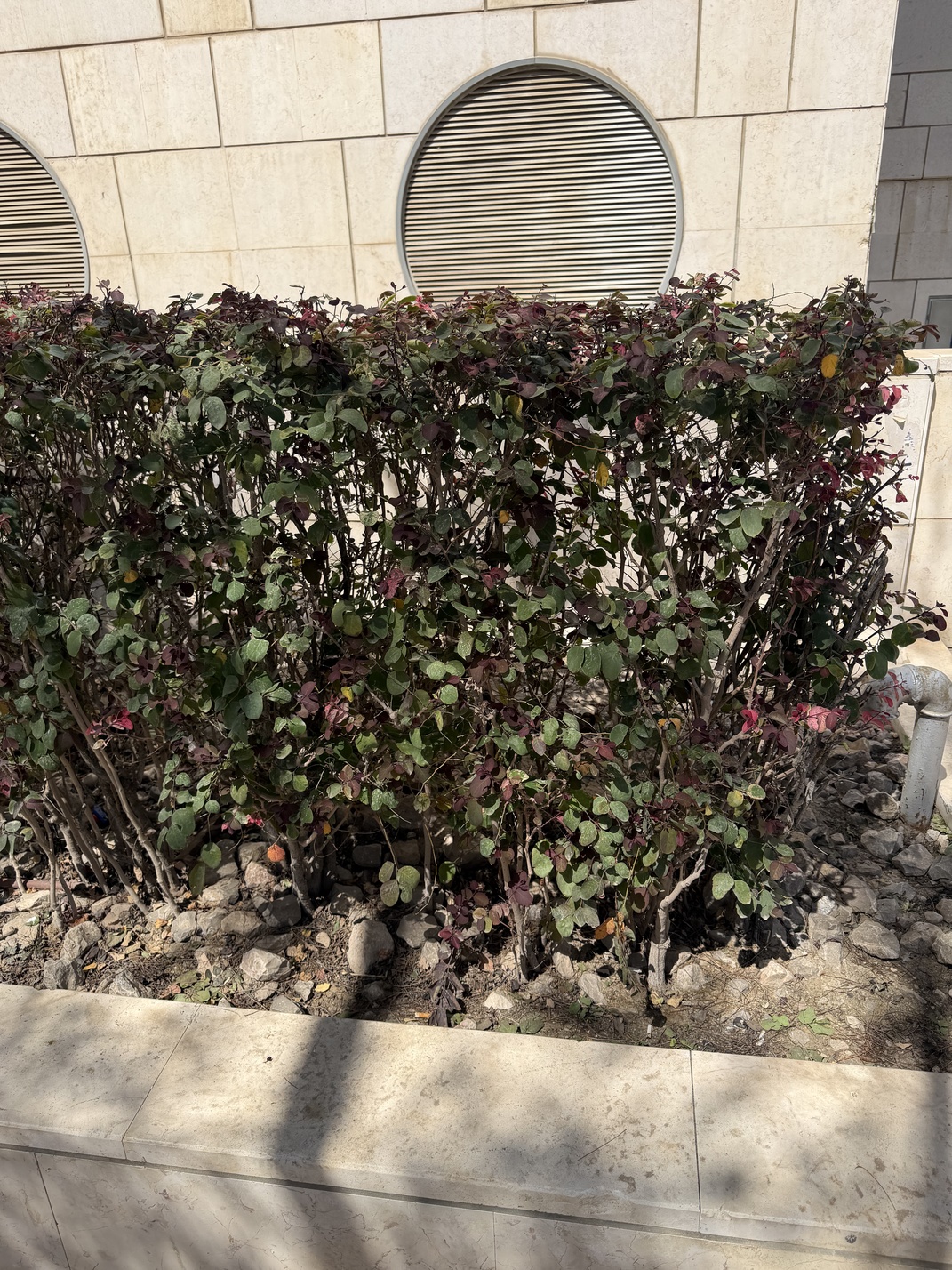} &
        \includegraphics[width=0.11\linewidth]{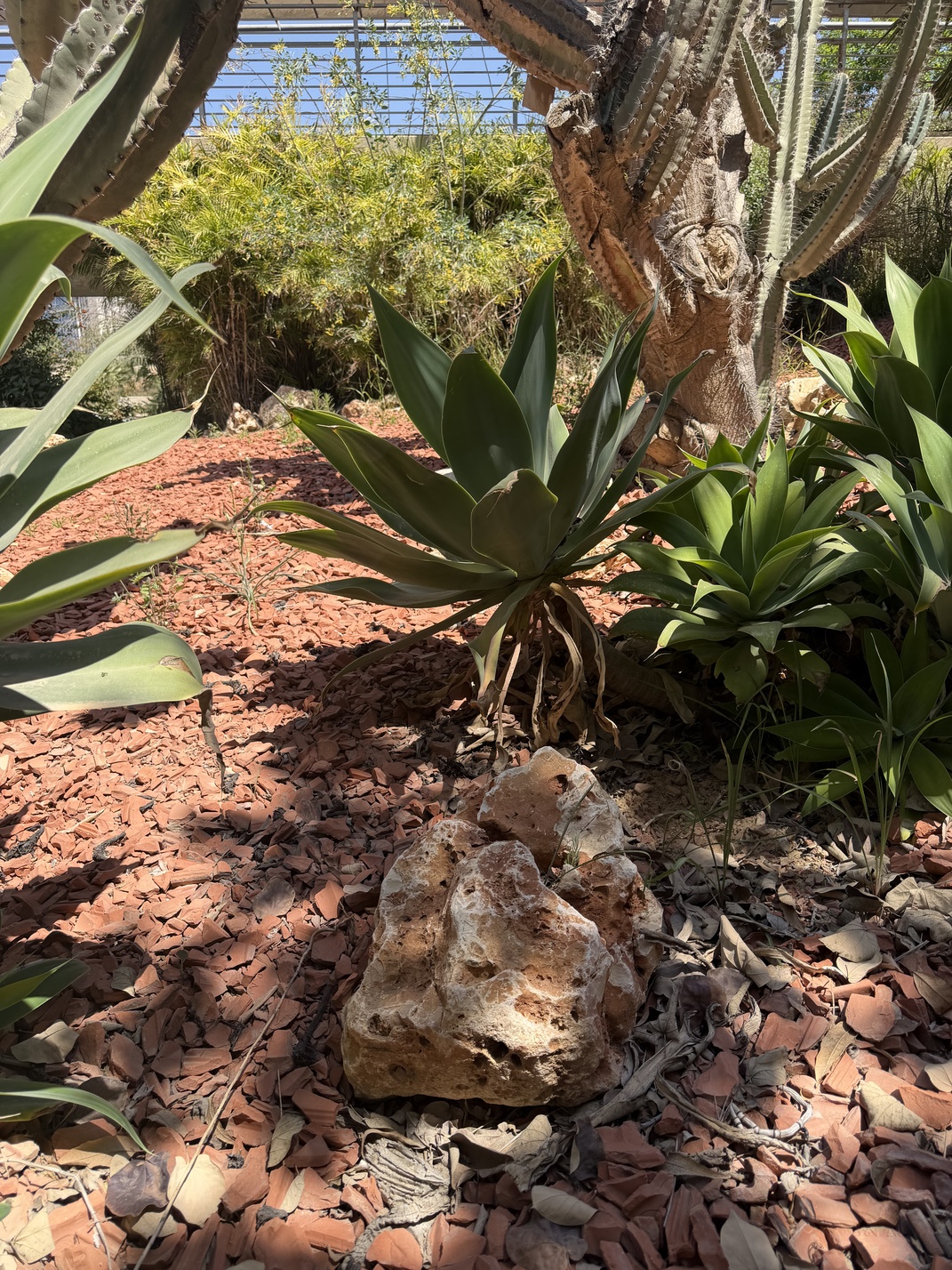} &
        \includegraphics[width=0.11\linewidth]{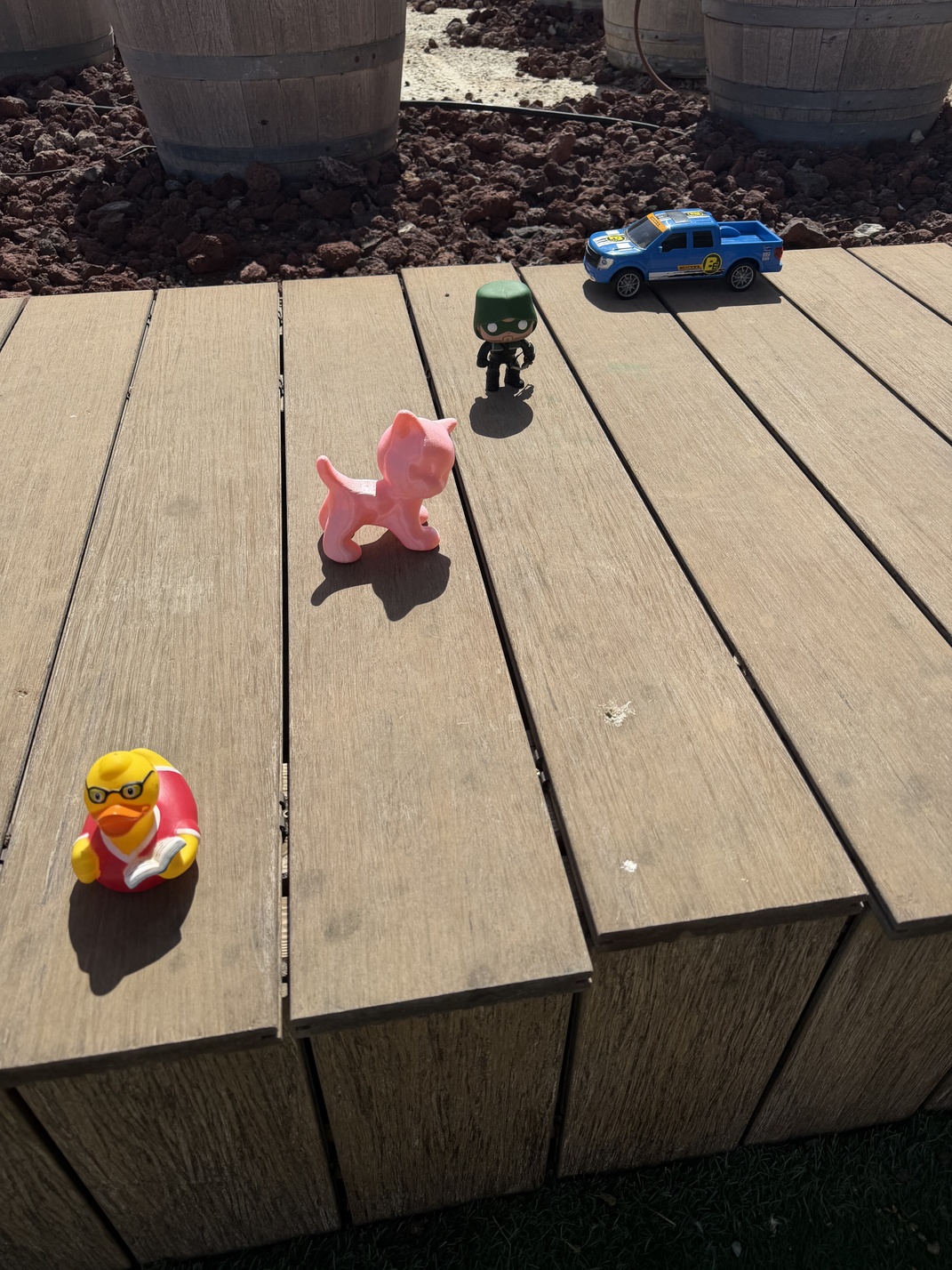} &
        \includegraphics[width=0.11\linewidth]{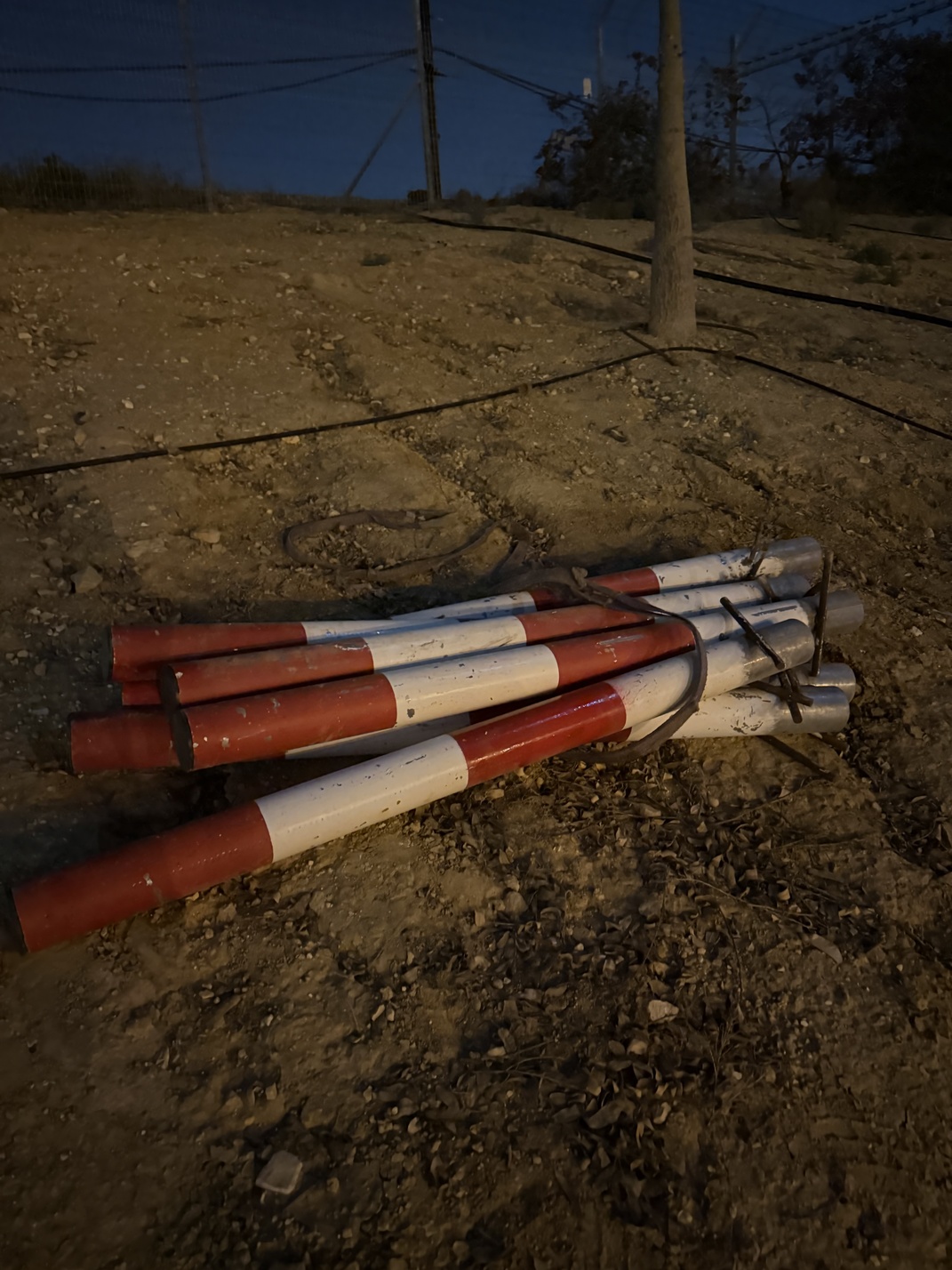} &
        \includegraphics[width=0.11\linewidth]{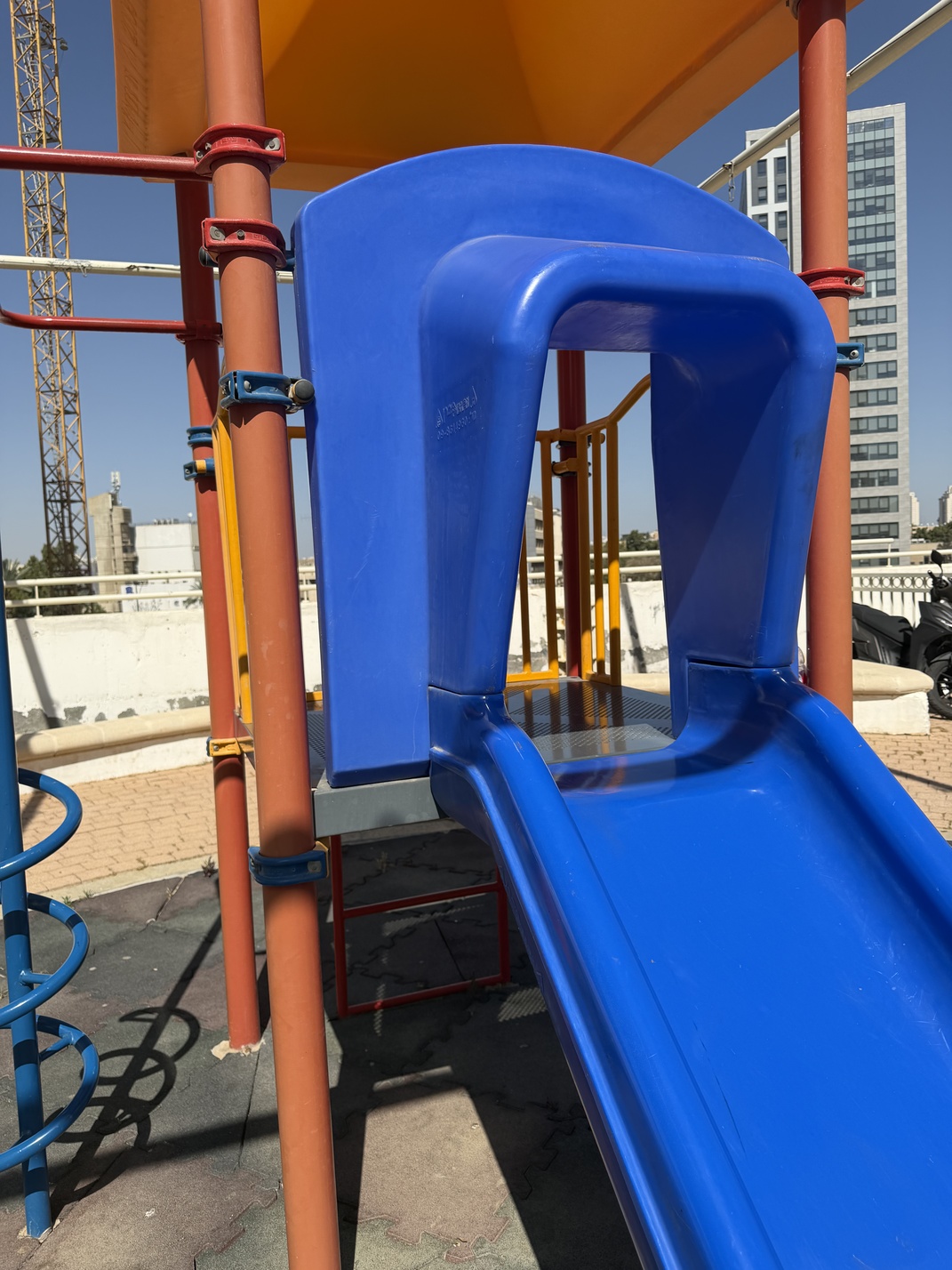} &
        \includegraphics[width=0.11\linewidth]{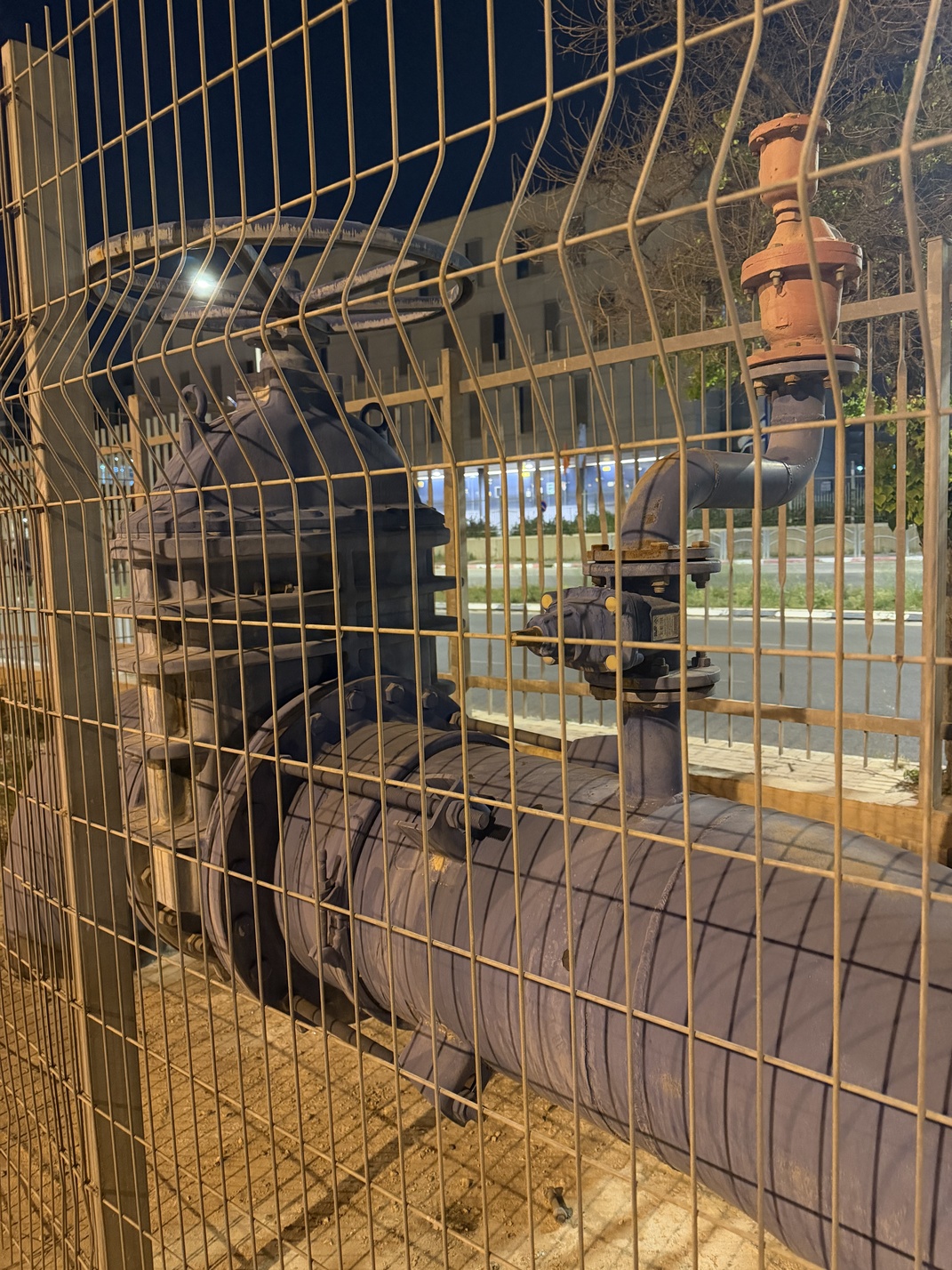} &
        \includegraphics[width=0.11\linewidth]{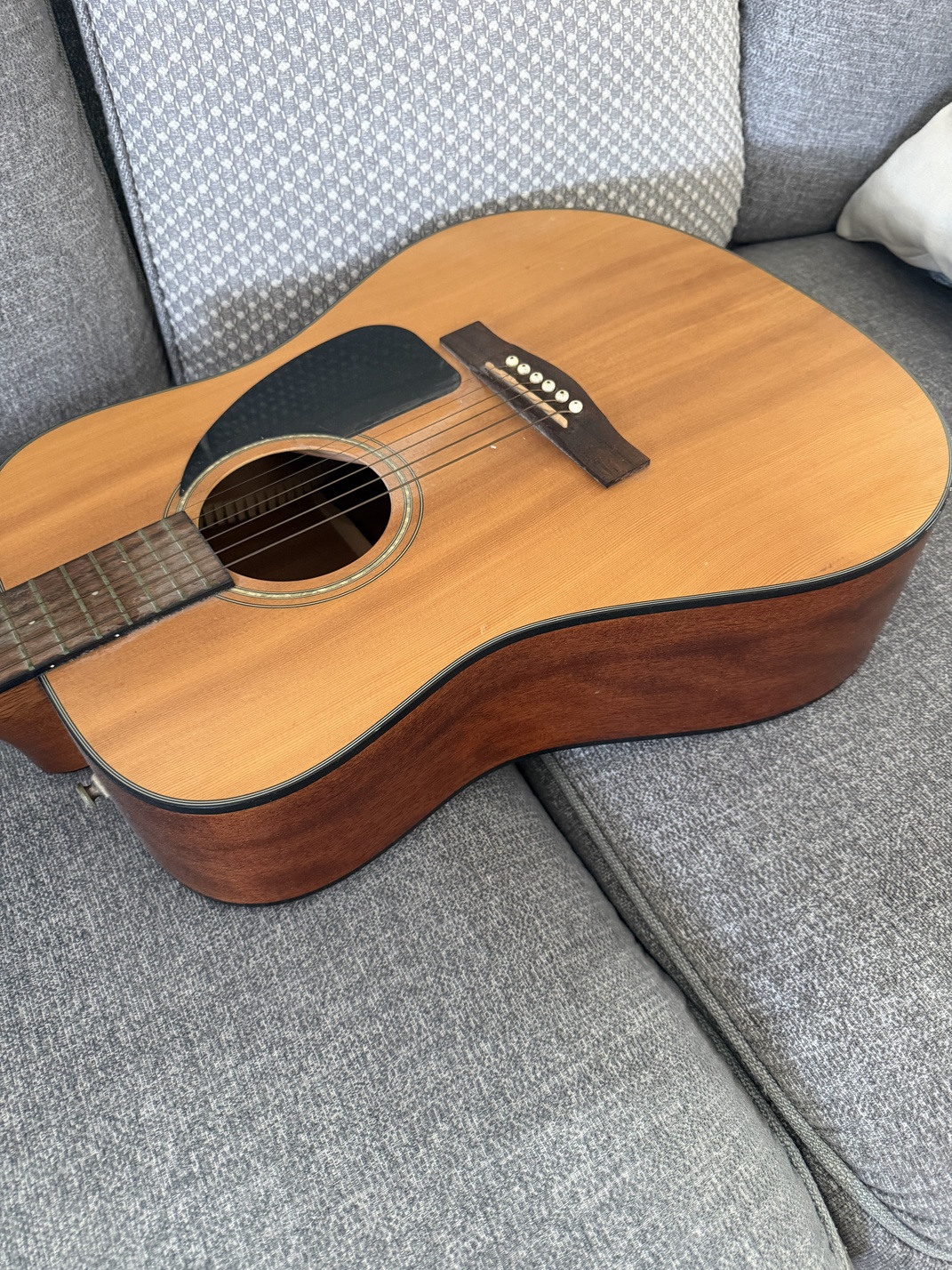}
    \end{tabular}%
    }
    \caption{\textbf{Example scenes from MobileClose-10.} The first row shows far training views, while the second row shows the close-up views. Each column represents examples from one scene.}
    \label{fig:self_collected_examples}
\end{figure*}

Each scene contains 15-18 far-view images used for training, together with 2-6 close-up evaluation views, resulting in a total of 39 close-up evaluation images. The close-up views are captured with significant changes in camera distance relative to the training views, with a median zoom factor of $3.42\times$ between the close-up evaluation views and the corresponding far views. Example far and close-up views from this benchmark are shown in Figure \ref{fig:self_collected_examples}.

This dataset is designed to emphasize challenging close-up novel view synthesis scenarios, where the target views exhibit substantial scale changes with the training views.

\end{document}